\documentclass[10pt,twocolumn,letterpaper]{article}

\usepackage[pagenumbers]{wacv}

\ifdefined\pdfobjcompresslevel \pdfobjcompresslevel=0 \fi
\usepackage{multirow}   
\usepackage{colortbl}   
\usepackage{tikz}       
\usepackage{microtype}  

\definecolor{wacvblue}{rgb}{0.21,0.49,0.74}
\usepackage[pagebackref,breaklinks,colorlinks,allcolors=wacvblue]{hyperref}

\def\wacvPaperID{2202} 
\def\confName{WACV}
\def\confYear{2027}

\title{GenAU: Language-Grounded Industrial Anomaly Understanding with Vision-Language Models}

\author{%
\begin{minipage}{\textwidth}\centering
\large
Hongkuan Zhou\textsuperscript{1,2} \quad
Tristan Rehm\textsuperscript{2} \quad
Nadeem Nazer\textsuperscript{3} \quad
Lavdim Halilaj\textsuperscript{1} \quad \\
Jingcheng Wu\textsuperscript{2} \quad
Steffen Staab\textsuperscript{2,4}\\[0.5em]
\textsuperscript{1}\,Corporate Research, Robert Bosch GmbH, Renningen, Germany \quad \\
\textsuperscript{2}\, Institute for Artificial Intelligence University of Stuttgart, Stuttgart, Germany \quad
\textsuperscript{3}\,Otto-von-Guericke-University Magdeburg, Magdeburg, Germany \quad \\
\textsuperscript{4}\,University of Southampton, Southampton, UK\\[0.4em]
\end{minipage}%
}

\begin{document}
\maketitle

\begin{abstract}
Industrial inspection requires more than binary anomaly detection: a practical system should determine whether an anomaly exists, localize the defective region, identify the defect type, and provide interpretable visual evidence. Existing CLIP-based methods detect and localize anomalies well but offer limited language-level defect understanding, while instruction-tuned vision-language models can describe defects but do not natively produce pixel-level masks. We introduce GenAU, a \textbf{Gen}eralist vision-language framework for industrial \textbf{A}nomaly \textbf{U}nderstanding that unifies image-level detection, pixel-level segmentation, multi-type anomaly detection, and defect analysis in a single instruction-following model. GenAU augments a vision-language model with two segmentation tokens, \texttt{[SEG\_defect]} and \texttt{[SEG\_normal]}, whose hidden states act as language-grounded queries over multi-scale visual features for pixel-level localization; the image-level score fuses this map with the decoder's textual normal/defect decision, while the language decoder produces structured defect-aware responses. Trained with a joint language-modeling and segmentation objective, GenAU covers all four tasks within one architecture and recipe, adding zero-shot multi-type detection and language-grounded defect analysis at a quantified cost to detection and segmentation. Across cross-dataset benchmarks, GenAU attains the strongest image-level detection among CLIP-based zero-shot methods on VisA and Real-IAD, with segmentation approaching but not surpassing specialized CLIP baselines.
\end{abstract}

\section{Introduction}
\label{sec:intro}
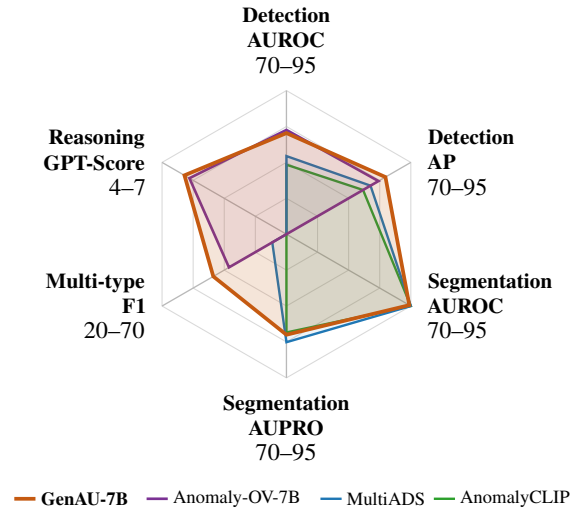
\begin{figure}[t]
    \centering
    \definecolor{cGenAU}{RGB}{197,90,17}   
    \definecolor{cAOV}{RGB}{123,50,148}    
    \definecolor{cMADS}{RGB}{31,119,180}   
    \definecolor{cCLIP}{RGB}{44,160,44}    
    \begin{tikzpicture}[scale=1.0]
      \def\R{1.9}            
      \foreach \f in {0.25,0.5,0.75,1.0}{
        \draw[gray!30,line width=0.4pt]
          (90:{\f*\R}) -- (30:{\f*\R}) -- (-30:{\f*\R}) --
          (-90:{\f*\R}) -- (-150:{\f*\R}) -- (150:{\f*\R})
          -- cycle;
      }
      \foreach \ang in {90,30,-30,-90,-150,150}{
        \draw[gray!55,line width=0.5pt] (0,0) -- (\ang:\R);
      }
      \newcommand{\radarpath}[6]{
        (90:{#1*\R}) -- (30:{#2*\R}) -- (-30:{#3*\R}) --
        (-90:{#4*\R}) -- (-150:{#5*\R}) -- (150:{#6*\R})
        -- cycle}
      \draw[cCLIP,line width=0.9pt,fill=cCLIP,fill opacity=0.08] \radarpath{0.484}{0.616}{1.0}{0.684}{0}{0};
      \draw[cMADS,line width=0.9pt,fill=cMADS,fill opacity=0.08] \radarpath{0.544}{0.676}{1.0}{0.752}{0.114}{0};
      \draw[cAOV,line width=1.0pt,fill=cAOV,fill opacity=0.08] \radarpath{0.726}{0.74}{0}{0}{0.462}{0.78};
      \draw[cGenAU,line width=1.4pt,fill=cGenAU,fill opacity=0.16]
        \radarpath{0.704}{0.796}{0.988}{0.700}{0.588}{0.82};
      \node[font=\footnotesize\bfseries,above=3pt,align=center,
            text width=2.5cm] at (90:\R) {Detection\\AUROC\\{\small\mdseries 70--95}};
      \node[font=\footnotesize\bfseries,anchor=west,xshift=3pt,align=left,
            text width=2.0cm] at (30:\R) {Detection\\AP\\{\small\mdseries 70--95}};
      \node[font=\footnotesize\bfseries,anchor=west,xshift=3pt,align=left,
            text width=2.2cm] at (-30:\R) {Segmentation\\AUROC\\{\small\mdseries 70--95}};
      \node[font=\footnotesize\bfseries,below=3pt,align=center,
            text width=2.5cm] at (-90:\R) {Segmentation\\AUPRO\\{\small\mdseries 70--95}};
      \node[font=\footnotesize\bfseries,anchor=east,xshift=-3pt,align=right,
            text width=1.9cm] at (-150:\R) {Multi-type\\F1\\{\small\mdseries 20--70}};
      \node[font=\footnotesize\bfseries,anchor=east,xshift=-3pt,align=right,
            text width=2.0cm] at (150:\R) {Reasoning\\GPT-Score\\{\small\mdseries 4--7}};
      \begin{scope}[shift={(-3.6,-3.5)},font=\scriptsize]
        \draw[cGenAU,line width=1.4pt] (0,0) -- (0.28,0)
          node[right,black,inner sep=1.5pt]{\textbf{GenAU-7B}};
        \draw[cAOV,line width=1.0pt] (1.75,0) -- (2.03,0)
          node[right,black,inner sep=1.5pt]{Anomaly-OV-7B};
        \draw[cMADS,line width=0.9pt] (4.05,0) -- (4.33,0)
          node[right,black,inner sep=1.5pt]{MultiADS};
        \draw[cCLIP,line width=0.9pt] (5.55,0) -- (5.83,0)
          node[right,black,inner sep=1.5pt]{AnomalyCLIP};
      \end{scope}
    \end{tikzpicture}
    \caption{Motivation of GenAU. Industrial inspection poses four questions per image (is there an anomaly, where, what type, and why), mapped to six metrics: image-level Detection (AUROC, AP), pixel-level Segmentation (AUROC, AUPRO), Multi-type recognition (F1), and Reasoning (GPT-Score). Each axis is scaled to the range beneath its label; a vertex at the center marks a task the model cannot perform (e.g., AnomalyCLIP and MultiADS emit no language; Anomaly-OV produces no segmentation). \textbf{GenAU is the only model in our study spanning all four}, though on segmentation it does not surpass the specialized CLIP baselines and on VisA image-level detection it trails the reasoning specialist Anomaly-OV. Detection is on VisA, Segmentation averaged over VisA/MPDD/Real-IAD, Multi-type over VisA/MPDD, and Reasoning the VisA-D\&R GPT-Score.}
    \label{fig:motivation}
\end{figure}
Industrial anomaly detection (AD) is fundamental to modern manufacturing, ensuring product quality, operational reliability, and safety. Automated visual inspection systems prevent defective products from reaching customers, thereby reducing financial losses, reputational risks, and safety hazards~\cite{effect-recall, HORA2011766}, while reducing reliance on human operators and supporting earlier intervention in production pipelines.
However, real-world industrial inspection involves more than simply deciding whether an image is normal or not. Instead, in practice, users often care about four closely related questions: 1) Is there an anomaly? 2) Where is it? 3) What type of anomaly is it? 4) Why are they anomalies? Therefore, a practical system should not only detect anomalies, but also localize them, recognize their types, and provide evidence for decisions. 

Despite this progress, most methods address only part of these needs. Recent zero-shot approaches~\cite{ad-review, ad-survery, zhou2024anomalyclip, chen2023zero, sadikaj2025multiadsdefectawaresupervisionmultitype} build on CLIP~\cite{radford2021} to distinguish normal from anomalous samples and derive anomaly maps, but remain focused on visual discrimination and localization, with limited support for semantic understanding; even multi-type extensions like MultiADS~\cite{sadikaj2025multiadsdefectawaresupervisionmultitype} do not explain \emph{why} a region is anomalous.

Large instruction-following VLMs (e.g., LLaVA~\cite{liu2023visualinstructiontuning}, Qwen-VL~\cite{bai2023qwenvlversatilevisionlanguagemodel}, and NVILA~\cite{liu2025nvilaefficientfrontiervisual}) offer a complementary direction with strong multimodal reasoning, but their potential for industrial anomaly understanding is underexplored. Existing VLM approaches such as AnomalyGPT~\cite{gu2023anomalygptdetectingindustrialanomalies} obtain localization from an external detector rather than the VLM itself, so prior methods emphasize either localization or language reasoning without tightly unifying detection, segmentation, type recognition, and explanation in one model.


To address this gap, we present GenAU, a Generalist vision-language framework for industrial Anomaly Understanding. GenAU unifies anomaly detection, anomaly segmentation, multi-type anomaly recognition, and defect analysis within a single instruction-following architecture and training recipe (cf. Figure~\ref{fig:motivation}). Specifically, we extend an instruction-following VLM with two dedicated segmentation tokens, \texttt{[SEG\_defect]} and \texttt{[SEG\_normal]}, which act as text-based anchors for defective and normal regions. By aligning their embeddings with multi-scale visual features from the visual encoder, GenAU produces fine-grained anomaly maps while preserving the model’s ability to recognize and describe specific defect types in natural language. In this way, GenAU combines the reasoning capabilities of large VLMs with pixel-level localization, moving beyond text-only defect recognition toward joint semantic understanding and spatial grounding.


We construct instruction-response data from standard benchmarks, training on MVTec-AD~\cite{Bergmann_2019_CVPR} and evaluating cross-dataset on VisA~\cite{VisA}, MPDD~\cite{mpdd}, and Real-IAD~\cite{wang2024realiadrealworldmultiviewdataset} under distribution shift. At inference GenAU is single-image, requiring no reference images or task-specific retraining.
Our contributions are summarized as follows:
\begin{itemize}
    \item We introduce GenAU, a generalist vision-language framework for industrial anomaly understanding. Through a unified instruction interface, GenAU supports image-level anomaly detection, pixel-level anomaly segmentation, multi-type anomaly detection, and textual defect analysis through a single instruction-following architecture and training recipe.

    \item We propose a language-grounded segmentation mechanism that extends an instruction-following VLM with dedicated \texttt{[SEG\_defect]} and \texttt{[SEG\_normal]} tokens. The hidden states of these tokens are aligned with multi-scale visual features, enabling the model to ground defect-aware language representations into pixel-level anomaly maps.

    \item We evaluate GenAU on public industrial anomaly benchmarks under cross-dataset distribution shift, covering binary anomaly detection, anomaly segmentation, and zero-shot multi-type anomaly detection. The results show that a single autoregressive VLM can jointly perform detection, segmentation, multi-type recognition, and defect reasoning (reaching the best image-level detection among CLIP-based zero-shot methods on VisA and Real-IAD), and we explicitly characterize the segmentation--reasoning trade-off incurred when the same recipe additionally learns to reason.
\end{itemize}

\section{Related Work}
\subsection{CLIP-based Prompt Adaptation for Anomaly Detection and Segmentation}
Most zero-shot industrial AD adapts CLIP~\cite{radford2021}, despite it not being designed for inspection. Prompt-based methods craft or learn normal/abnormal prompts: WinCLIP~\cite{jeong2023winclip} (window aggregation), AnomalyCLIP~\cite{zhou2024anomalyclip} (object-agnostic prompts), and PromptAD~\cite{Li_2024_WACV}/AdaCLIP~\cite{Cao_2024} (learnable prompt tokens). For localization, APRIL-GAN~\cite{chen2023zero}, AnoCLIP~\cite{deng2024}, and SimCLIP~\cite{simclip} improve patch-level visual--text alignment; CLIP-SAM~\cite{li2024clipsamclipsamcollaboration} and FiLo~\cite{gu2024filozeroshotanomalydetection} add external segmentation/grounding modules; and MultiADS~\cite{sadikaj2025multiadsdefectawaresupervisionmultitype} adds defect-aware supervision for multi-type segmentation, while DAPO~\cite{nazer2025dapo} learns hybrid defect-aware prompts via progressive tuning for multi-type detection and localization under distribution shift. Others rely on patch comparison or reconstruction~\cite{li2024musczeroshotindustrialanomaly,schwartz2024maeday,aota2023zero}. These methods largely depend on CLIP-style embedding comparison or external modules, making them less suited to unified defect reasoning and analysis.

\subsection{Autoregressive VLMs for Anomaly Detection and Reasoning}
Autoregressive VLMs offer a different paradigm, enabling open-ended visual reasoning and instruction following instead of fixed-prompt embedding comparison. AnomalyGPT~\cite{gu2023anomalygptdetectingindustrialanomalies} is an early effort, but it relies on an external CLIP-based anomaly map combined with the input image and fed to the LLM, so the VLM mainly serves as a reasoning interface over a separately generated map. More recently, Anomaly-OV~\cite{xu2025towards} specializes MLLMs for zero-shot anomaly detection and reasoning via the Anomaly-Instruct-125k instruction set, but targets image-level detection and explanation, with pixel-level segmentation not grounded in the decoding process.

In contrast, GenAU treats the autoregressive VLM as the core anomaly understanding model rather than as a reasoning interface over externally generated anomaly maps. Building on the reasoning-segmentation paradigm of LISA~\cite{lai2024lisareasoningsegmentationlarge}, in which a VLM emits a segmentation token whose hidden state drives a mask, GenAU uses decoder hidden states of dedicated \texttt{[SEG\_normal]} and \texttt{[SEG\_defect]} tokens as language-grounded anchors, aligning them directly with frozen multi-scale visual features (without an external mask decoder) to generate pixel-level anomaly maps. This design unifies image-level anomaly detection, anomaly segmentation, multi-type anomaly detection, and textual defect analysis in a single instruction-following framework. Grounding visual features in language is consistent with broader evidence that aligning visual representations with language or structured knowledge improves generalization under distribution shift~\cite{zhou2024kgv} and open-domain visual recognition~\cite{zhou2026knowcol}. An extended discussion is provided in Appendix.

\section{Preliminaries}
\label{sec:prelims}
We address three tasks under a common cross-dataset protocol. A model is trained only on a source dataset $\mathcal{D}_{\text{train}}=\{(\boldsymbol{x}_i,\boldsymbol{y}_i)\}_{i=1}^{N_1}$ with inputs $\boldsymbol{x}_i\sim\mathcal{P}(\mathcal{X})$, and evaluated, without retraining, on a target dataset $\mathcal{D}_{\text{target}}$ whose inputs are drawn from a different distribution $\mathcal{P}(\mathcal{X}')\nsim\mathcal{P}(\mathcal{X})$ (a distribution shift). At test time the model may optionally use target-domain metadata such as a product description, but no target labels. What the literature calls \emph{zero-shot} anomaly detection/segmentation is, more precisely, binary detection/segmentation under this category-level distribution shift; we use the terms interchangeably and retain \emph{zero-shot} for comparability with prior work.

\smallskip\noindent\textbf{Binary anomaly detection.} Each image $\boldsymbol{x}\in\mathbb{R}^{H\times W}$ carries a label $y\in\{0,1\}$ indicating anomaly presence; the model classifies each target image as normal or anomalous.

\smallskip\noindent\textbf{Binary anomaly segmentation.} Each image is paired with a pixel mask $\boldsymbol{y}\in\{0,1\}^{H\times W}$; the model classifies each target pixel as normal or anomalous.

\smallskip\noindent\textbf{Zero-shot multi-type detection.} Each image carries a multi-label vector over one normal class and $K$ defect types $\{d_1,\dots,d_K\}$. The target defect vocabulary partially overlaps the training one, $C_{\text{train}}\cap C_{\text{target}}\neq\emptyset$ and $C_{\text{target}}\setminus C_{\text{train}}\neq\emptyset$, so the target mixes seen and novel defect types. The model predicts $\hat{\boldsymbol{y}}'\in\{0,1\}^{K_2+1}$ over the target vocabulary. Full formal definitions are given in Appendix~\ref{app:tasks}.

\section{Methodology}
We present an overview of the proposed framework, followed by detailed descriptions of the anomaly detection and anomaly segmentation components, the training objectives, and the inference process. 

\subsection{Framework Overview}
\begin{figure*}[t]
  \centering
  \includegraphics[width=1.0\textwidth]{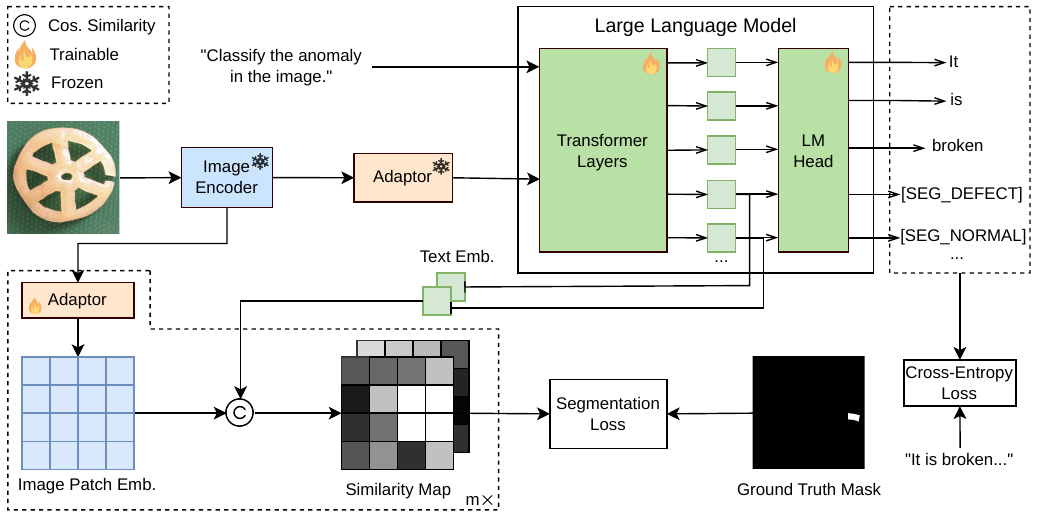}
  \caption[Overview of the training phase of GenAU.]{Overview of the training phase of GenAU. The model architecture processes visual information through two distinct pathways for anomaly classification and segmentation. First, the query image is processed by a frozen image encoder. The features from the final encoder layer are projected into the LLM's embedding space and combined with the prompt's token embeddings. This combined input is fed to the LLM, which is trained to generate the defect type and specialized \texttt{[SEG]} tokens. This classification path is supervised using a cross-entropy loss. Concurrently, image patch features from the intermediate and final encoder layers are routed through separate projection heads (one per layer). The projected visual features are compared with the \texttt{[SEG]} token’s hidden state from the last transformer layer (pre-LM head) to produce a similarity map. This segmentation path is supervised using a combination of focal, Tversky, and Dice losses.}
   \label{fig:Model-Arch}
\end{figure*}

GenAU couples two jointly-trained pathways (Figure~\ref{fig:Model-Arch}): (1) a \emph{text} pathway that generates text conversation and (2) a \emph{visual grounding} pathway that uses the hidden states of the generated \texttt{[SEG\_normal]}/\texttt{[SEG\_defect]} tokens as language-grounded anchors, aligned with multi-scale visual features to produce pixel-level anomaly maps; the image-level score fuses the map maximum with the decoder's normal/defect decision.

\subsection{Anomaly Segmentation}
\label{subsec:seg}
GenAU performs anomaly localization by grounding visual features with the hidden states of generated segmentation tokens. Given an image $\mathbf{X}_v$ and an instruction $\mathbf{X}_q$, the autoregressive decoder generates a structured response containing the special tokens \texttt{[SEG\_normal]} and \texttt{[SEG\_defect]}. Let $\mathbf{h}_{\mathrm{normal}}, \mathbf{h}_{\mathrm{defect}}\in\mathbb{R}^{D}$ denote their final-layer decoder hidden states. These two vectors serve as language-grounded anchors for the normal and defective states:
\begin{equation}
    \mathbf{T}
    =
    [\mathbf{h}_{\mathrm{normal}}, \mathbf{h}_{\mathrm{defect}}]
    \in \mathbb{R}^{2\times D}.
\end{equation}

The input image is processed by the frozen vision encoder to obtain multi-scale patch features from intermediate and final stages,
\(\{\mathbf{V}_i\}_{i=1}^{m}\), where
\(\mathbf{V}_i\in\mathbb{R}^{H_iW_i\times C_i}\) denotes the patch features from the \(i\)-th stage. Since the visual features and decoder hidden states lie in different representation spaces, each stage is equipped with its own \emph{linear} projection layer
\(\phi_i:\mathbb{R}^{C_i}\rightarrow\mathbb{R}^{D}\), producing
\begin{equation}
    \mathbf{Z}_i
    =
    \phi_i(\mathbf{V}_i)
    \in
    \mathbb{R}^{H_iW_i\times D}.
\end{equation}

For each stage, we compute cosine-similarity logits between every projected patch feature and the two segmentation-token anchors:
\begin{equation}
    \mathbf{S}_i
    =
    \gamma
    \cdot
    \frac{\mathbf{Z}_i}{\|\mathbf{Z}_i\|_2}
    \left(
    \frac{\mathbf{T}}{\|\mathbf{T}\|_2}
    \right)^{\top},
    \qquad
    \mathbf{S}_i\in\mathbb{R}^{H_iW_i\times 2},
\end{equation}
where \(\gamma\) is a fixed temperature scale. The two channels of \(\mathbf{S}_i\) correspond to the normal and defect scores. After reshaping \(\mathbf{S}_i\) to spatial resolution \(H_i\times W_i\), the maps are upsampled to the target mask resolution and normalized with a channel-wise softmax. The defect-channel probability gives the pixel-level anomaly map:
\begin{equation}
    \mathbf{A}_i
    =
    \mathrm{Softmax}(\mathrm{UP}(\mathbf{S}_i))_{[\mathrm{defect}]}
    \in
    [0,1]^{H\times W}.
\end{equation}
The stage-wise maps are aggregated by summation across the $m$ stages to form the final anomaly map $\mathbf{A}$, where higher values indicate a higher probability that the corresponding pixels belong to defective regions.

\paragraph{AnyRes high-resolution readout.}
The backbone encodes high-resolution inputs with an AnyRes scheme that, besides a global low-resolution view, splits the image into a grid of higher-resolution tiles. Applying the readout above to the global view alone yields a coarse low-resolution grid. To recover fine spatial detail, GenAU reassembles the patch features of the AnyRes tiles into a single high-resolution feature grid and computes the same anchor-grounded similarity maps on it. At inference, the high-resolution map $\mathbf{A}^{\mathrm{hi}}$ and the global-view map $\mathbf{A}^{\mathrm{g}}$ are fused per stage as $\mathbf{A}=(1-\beta)\,\mathbf{A}^{\mathrm{g}}+\beta\,\mathbf{A}^{\mathrm{hi}}$ with a fixed weight $\beta$. This high-resolution readout is used for all reported segmentation results.

\subsection{Image-level Anomaly Scoring}
Building on the segmentation map above, GenAU computes the image-level anomaly score without training a separate image-level classifier. Let $\mathbf{A}$ denote the final anomaly map of Section~\ref{subsec:seg} (the stage-aggregated defect response). The image-level anomaly score is its spatial maximum,
\begin{equation}
    s_{\mathrm{img}}
    =
    \max_{(u,v)}\mathbf{A}(u,v),
\end{equation}
where a larger $s_{\mathrm{img}}$ indicates a higher probability that the image contains an anomaly. This score introduces no additional trainable parameters and is obtained entirely from the language-grounded segmentation readout.

\paragraph{Fused image-level score.}
The map maximum $s_{\mathrm{img}}$ is a purely visual readout. Because the decoder additionally emits an explicit normal/defect decision, the image-level score we report is
\begin{equation}
    s_{\mathrm{fused}}
    =
    \tfrac{1}{2}\,\tilde{s}_{\mathrm{img}}
    +
    \tfrac{1}{2}\,\tilde{\ell}_{\mathrm{nd}},
\end{equation}
where the visual score $\tilde{s}_{\mathrm{img}}$ and the decoder's normal/defect log-odds $\tilde{\ell}_{\mathrm{nd}}$ are each standardized by a frozen calibration estimated once on MVTec (Appendix~\ref{app:impl}); no target-domain labels are used. Unless stated otherwise, all reported image-level results use $s_{\mathrm{fused}}$.

\subsection{Textual Reasoning and Multi-Type Anomaly Detection}
For zero-shot multi-type anomaly detection, given a target image $\mathbf{X}_v$, a language instruction $\mathbf{X}_q$, and the target-domain defect vocabulary $C_{\mathrm{target}}=\{d'_1,\ldots,d'_{K_2}\}$, GenAU generates a structured response $\mathbf{X}_a$ with an autoregressive VLM. A deterministic parser $\pi(\cdot)$ maps the response to a multi-type anomaly label $\hat{\mathbf{y}}'\in\{0,1\}^{K_2+1}$, where the entries correspond to the normal class and the $K_2$ target defect types. 

\subsection{Training Objective}

GenAU is optimized with a joint objective that supervises textual reasoning and pixel-level anomaly segmentation. For each training sample, the model receives an image $\mathbf{X}_v$ and an instruction $\mathbf{X}_q$, and is trained to generate a structured response $\mathbf{X}_a=(a_1,\ldots,a_T)$ containing the predicted defect type and the grounding tokens \texttt{[SEG\_normal]} and \texttt{[SEG\_defect]}. The overall objective is
\begin{equation}
    \mathcal{L}
    =
    \mathcal{L}_{\mathrm{txt}}
    +
    \lambda_{\mathrm{seg}}\mathcal{L}_{\mathrm{seg}},
    \label{eq:overall-loss}
\end{equation}
where $\lambda_{\mathrm{seg}}$ balances the language-modeling and pixel-level supervision terms. The image-level anomaly score is read off the segmentation map at inference and thus requires no dedicated training term.

\paragraph{Text generation loss.}
The textual reasoning objective is the standard autoregressive cross-entropy loss over the target response tokens:
\begin{equation}
    \mathcal{L}_{\mathrm{txt}}
    =
    -
    \sum_{t=1}^{T}
    \log
    p_{\theta}
    \left(
        a_t
        \mid
        a_{<t}, \mathbf{X}_v, \mathbf{X}_q
    \right).
    \label{eq:text-loss}
\end{equation}
This term trains the model to generate domain-specific inspection responses, including defect-type descriptions and the required segmentation tokens. Prompt tokens and inserted visual tokens are excluded from the language modeling loss.

\paragraph{Pixel-level segmentation loss.}
For anomaly segmentation, GenAU predicts stage-wise similarity maps $\{\mathbf{S}_i\}_{i=1}^{m}$, where each $\mathbf{S}_i$ contains normal and defect logits for visual patches at stage $i$. Let $\mathbf{A}_i$ and $\mathbf{A}_i^{\mathrm{n}}$ denote the defect- and normal-channel probability maps after upsampling and channel-wise softmax. Given a binary ground-truth mask $\mathbf{M}_x\in\{0,1\}^{H\times W}$, the segmentation loss is summed over all visual stages:
\begin{align}
    \mathcal{L}_{\mathrm{seg}}
    =
    \sum_{i=1}^{m}
    \Big(
        &
        \mathcal{L}_{\mathrm{focal}}
        \big(
            \mathrm{UP}(\mathbf{S}_i),
            \mathbf{M}_x
        \big)
        +
        \mathcal{L}_{\mathrm{Tversky}}
        \big(
            \mathbf{A}_i,
            \mathbf{M}_x
        \big)
        \notag \\
        &
        +
        \mathcal{L}_{\mathrm{dice}}
        \big(
            \mathbf{A}_i^{\mathrm{n}},
            \mathbf{1}-\mathbf{M}_x
        \big)
    \Big).
    \label{eq:seg-loss}
\end{align}
The focal term mitigates the severe foreground/background imbalance; the Tversky term on the defect channel up-weights false negatives over false positives, improving region recall and AUPRO; and the Dice term on the normal channel suppresses false positives. For anomalous images $\mathbf{M}_x$ is the annotated mask; for normal images it is all-zero, giving negative supervision.

\paragraph{Trainable parameters.}
We use parameter-efficient fine-tuning: the vision encoder and cross-modal projector stay frozen, while LoRA~\cite{hu2021loralowrankadaptationlarge} on the decoder, the token-embedding table and LM head (for the two segmentation tokens), the $m$ stage-wise projectors $\{\phi_i\}$, and the decoder layer-norms are trained jointly under Eq.~\ref{eq:overall-loss}.

\subsection{Inference}
GenAU uses the same instruction format as in training. For binary detection and segmentation the instruction names the object category and asks, via natural-language paraphrases, \texttt{``Is there any anomaly on the \{object\}? If yes, segment it.''} For zero-shot multi-type detection it is instead conditioned on the product type and lists the target object's candidate defect types (its coarse classes under the taxonomy of Appendix~\ref{app:taxonomy}), so different products receive different option sets: \texttt{``Identify the defect type. Choose exactly one option from this list: \{defect list\}.''}

The model autoregressively generates a textual answer immediately followed by the two segmentation tokens, e.g.\ \texttt{\{defect\} [SEG\_defect] [SEG\_normal]} for an anomalous image, and a statement of normality followed by the same tokens for a normal one. Crucially, both \texttt{[SEG\_normal]} and \texttt{[SEG\_defect]} are \emph{always} emitted, so their final-layer hidden states are available as language-grounded anchors regardless of the textual prediction. From these anchors we read off the three task outputs: (i)~binary detection thresholds the fused image-level score $s_{\mathrm{fused}}$, while threshold-free metrics such as image-level AUROC are computed directly from it; (ii)~segmentation forms the AnyRes-fused defect-probability map of Sec.~\ref{subsec:seg}, where high responses mark defective regions and normal images are expected to yield low maps everywhere (they are supervised with all-zero masks during training); and (iii)~a deterministic parser $\pi(\cdot)$ maps the multi-type response to $\hat{\mathbf{y}}'\in\{0,1\}^{K_2+1}$ over the normal class and the $K_2$ target defect types.

\section{Experiments}
In this section, we describe the experimental setup, datasets, and evaluation metrics used to assess the performance of GenAU. We also present the baseline methods and key implementation details.

\subsection{Datasets}
We use four public industrial anomaly benchmarks (MVTec-AD~\cite{Bergmann_2019_CVPR}, VisA~\cite{VisA}, MPDD~\cite{mpdd}, and Real-IAD~\cite{wang2024realiadrealworldmultiviewdataset}), plus VisA-D\&R~\cite{xu2025towards} for defect reasoning, covering the three tasks of Sec.~\ref{sec:prelims} and defect reasoning/analysis. Additional details are in Appendix.

\subsection{Experimental Settings}
We build GenAU on LLaVA-OneVision~\cite{li2025llavaonevision}, which pairs a Qwen2 language backbone with a frozen SigLIP-so400m-patch14-384 vision encoder, and report results at the 0.5B and 7B scales. Following prior zero-shot protocols~\cite{sadikaj2025multiadsdefectawaresupervisionmultitype, chen2023zero, zhou2024anomalyclip}, GenAU is fine-tuned on the test split of MVTec-AD (its only split carrying anomaly masks and defect-type labels, the train split being normal-only) and evaluated, without retraining, on VisA, MPDD, and Real-IAD; MVTec-AD is thus purely the source domain and is disjoint from every evaluation target.
High-resolution inputs use AnyRes (\texttt{anyres\_max\_9}, up to a $6\times6$ grid of $384$-px tiles); the segmentation readout reassembles the tiles into a high-resolution feature grid, fused with the global view ($\beta=0.25$, temperature $20$) using patch features from vision-encoder layers $6,12,18,27$. We adapt only LoRA~\cite{hu2021loralowrankadaptationlarge} ($r=16$, $\alpha=32$) on the LLM, the token embeddings and LM head (to add the two segmentation tokens), and the per-stage projectors and decoder layer-norms; the encoder and cross-modal projector stay frozen. The segmentation loss combines focal, Tversky ($\alpha=0.3$, $\beta=0.7$), and Dice, with $\lambda=1$ (Eq.~\ref{eq:overall-loss}) and no separate image-level loss; the image score fuses the defect-map maximum with the decoder log-odds via a frozen MVTec calibration (mixing weight $0.5$). We train one epoch at effective batch size $32$ ($lr=10^{-4}$ cosine, warmup $0.03$, no weight decay; $8$ GPUs, ZeRO-3). Further details, including the multi-type prompt and parser, are in Appendix.

\subsection{Evaluation Metrics}
We report image-level AUROC and AP for detection; pixel-level AUROC and AUPRO for segmentation; weighted Precision/Recall/F1 for multi-type detection (to handle class imbalance); and, for reasoning, the VisA-D\&R GPT-Score~\cite{xu2025towards}, where GPT-4o rates each response $1$--$10$ for low-level (perception) and complex (causal-reasoning) questions separately (full definitions in Appendix).

\subsection{Baselines}
All baselines, like GenAU, access only the query image at test time: no reference, support, or exemplar images. As \emph{CLIP-based} zero-shot detectors and segmenters we compare against CLIP, CLIP-AC~\cite{radford2021}, WinCLIP~\cite{jeong2023winclip}, APRIL-GAN~\cite{chen2023zero}, CoOp~\cite{CoOp}, CoCoOp~\cite{zhou2022conditionalpromptlearningvisionlanguage}, AnomalyCLIP~\cite{zhou2024anomalyclip}, AdaCLIP~\cite{Cao_2024}, FiLo~\cite{gu2024filozeroshotanomalydetection}, and MultiADS together with its variant MultiADS-F~\cite{sadikaj2025multiadsdefectawaresupervisionmultitype}, the only CLIP-based baseline that additionally performs multi-type detection. For the multi-type and reasoning tasks we further compare against \emph{vision--language} baselines: the anomaly-reasoning specialist Anomaly-OV~\cite{xu2025towards} and our LLaVA-OneVision~\cite{li2025llavaonevision} backbone, both at the $0.5$B and $7$B scales. Methods without public implementations (e.g., SimCLIP~\cite{simclip}) are excluded for reproducible comparison.

\subsection{Results}
Next, we present and discuss results for binary anomaly detection and segmentation under distribution shift, multi-type zero-shot anomaly detection, and defect reasoning, the four tasks GenAU addresses within a single architecture and training recipe.
\subsubsection{Binary Detection and Segmentation under Distribution
Shift}
Table~\ref{tab:binary} reports the performance of GenAU on VisA, MPDD, and Real-IAD, compared with state-of-the-art CLIP-based methods. Pixel-level metrics (AUROC, AUPRO) and image-level metrics (AUROC, AP) are reported.

\begin{figure}[h]
  \centering
  \includegraphics[width=1.0\columnwidth]{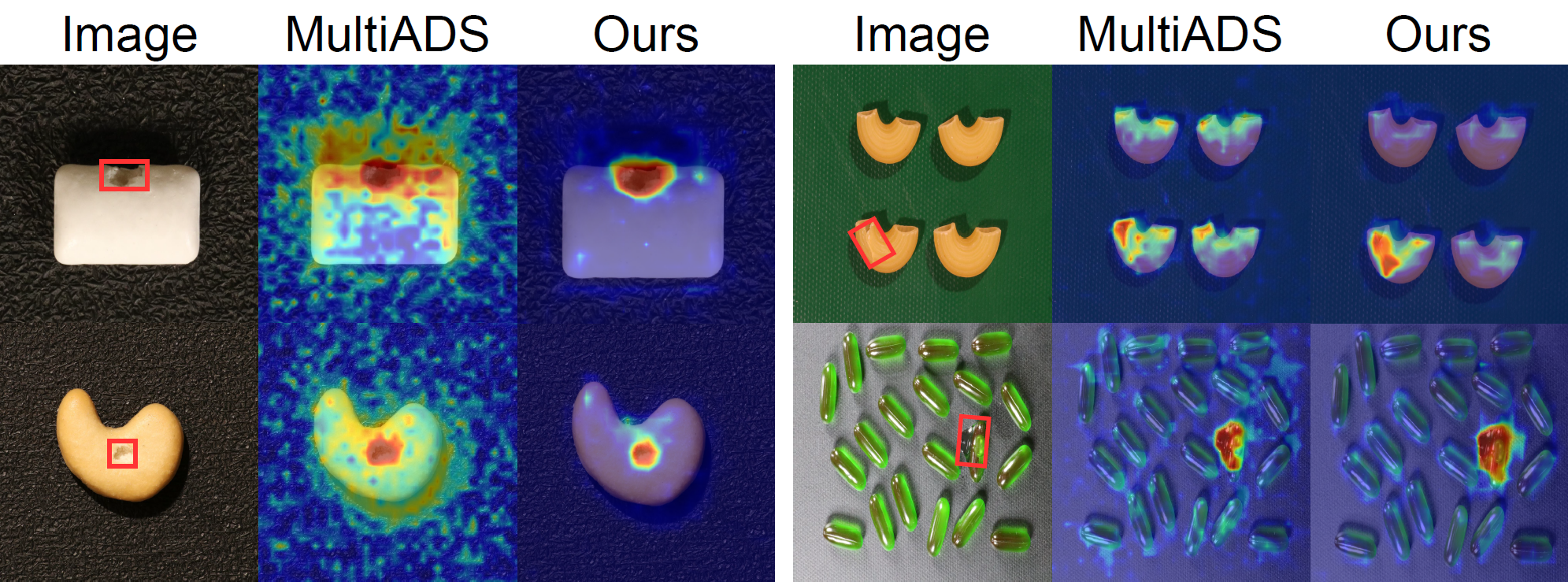}
  \caption{Visualization of anomaly segmentation under distribution shift from VisA. Results are compared with MultiADS.}
   \label{fig:visa-visual-good}
\end{figure}

On VisA, the 7B model achieves the best image-level AUROC ($87.6\%$) among the CLIP-based zero-shot methods in Table~\ref{tab:binary}, together with competitive pixel-level scores ($94.7\%$ AUROC); the VLM reasoning specialist Anomaly-OV-7B reports a higher VisA image-AUROC ($88.9\%$) but is trained under an instruction-tuning rather than our cross-dataset protocol and produces no segmentation, so we note it here only for context. We note, further, that GenAU's pixel-level AUROC does not surpass the strongest CLIP-based segmenters on any of the three datasets. The 0.5B variant remains competitive. On MPDD, the 7B model attains high pixel-level AUROC/AUPRO but lower image-level AUROC ($70.3\%$, against $79.7\%$ for MultiADS-F), reflecting the logical and large-area anomalies characteristic of this dataset. On Real-IAD, the 7B model reaches an image-level AUROC of $84.7\%$. Across datasets, pixel-level AUPRO is consistently the lowest-scoring metric.
Overall, GenAU demonstrates that a single instruction-following VLM can achieve strong binary detection and segmentation performance under distribution shift.
\begin{table}
\setlength{\tabcolsep}{1pt}
\centering
	\caption[Results of GenAU for binary anomaly detection and segmentation under distribution shift]{Results for binary anomaly detection and segmentation under distribution shift of GenAU and baselines. All metrics are in \%. \textbf{Bold} represents the best performer and \underline{underline} indicates the second best performer. Baseline results are taken from the  MultiADS~\cite{sadikaj2025multiadsdefectawaresupervisionmultitype} and FiLo~\cite{gu2024filozeroshotanomalydetection} papers and are not recomputed in our harness.}
	\label{tab:binary}
\begin{tabular}{c|ccccc}
\toprule \small
\multirow{2}{*}{\textbf{Dataset}} & \multirow{2}{*}{\textbf{Method}}  
& \multicolumn{2}{c}{\textbf{Pixel-Level}} & \multicolumn{2}{c}{\textbf{Image-Level}} \\
\cmidrule{3-6}
& & \textbf{AUROC} & \textbf{AUPRO} & \textbf{AUROC} & \textbf{AP} \\
\midrule
\multirow{13}{*}{VisA}
& CLIP        & 46.6 & 14.8 & 66.4 & 71.5 \\
& CLIP-AC     & 47.8 & 17.3 & 65.0 & 70.1 \\
& CoOp        & 24.2 & 3.8 & 62.8 & 68.1 \\
& CoCoOp      & 93.6 & - & 78.1 & - \\
& WinCLIP     & 79.6 & 56.8 & 78.1 & 81.2 \\
& APRIL-GAN   & 94.2 & 86.8 & 78.0 & 81.4 \\
& AnomalyCLIP  & \underline{95.5} & 87.0 & 82.1 & 85.4 \\
& AdaCLIP     & 95.0 & - & 75.4 & 79.3 \\
& FiLo     & \textbf{95.9} & - & \underline{83.9} & - \\
& MultiADS   & 95.0 & \textbf{89.7} & 83.6 & \underline{86.9}\\
& MultiADS-F & 94.5 & 87.4 & 82.5 & 86.5 \\
\cmidrule(lr){2-6}
&  GenAU-0.5b &      91.7 & 83.2 & 79.6 & 85.0 \\
&  GenAU-7b &      94.7 & \underline{88.2} & \textbf{87.6} & \textbf{89.9} \\
\midrule
\multirow{12}{*}{MPDD}
& CLIP        & 62.1 & 33.0 & 54.3 & 65.4 \\
& CLIP-AC     & 58.7 & 29.1 & 56.2 & 66.0 \\
& CoOp        & 15.4 & 2.3  & 55.1 & 64.2 \\
& CoCoOp      & 95.2 & - & 61.0 & - \\
& WinCLIP      & 76.4 & 48.9 & 63.6 & 69.9 \\
& APRIL-GAN    & 94.1 & 83.2 & 73.0 & 80.2 \\
& AnomalyCLIP  & \textbf{96.5} & 88.7 & 77.0 & \textbf{82.0} \\
& AdaCLIP      & \underline{96.3} & - & 66.3 & 75.0 \\
& MultiADS   & 95.8 & \textbf{89.7} &  \underline{78.3} & 78.4 \\
& MultiADS-F & \underline{96.3} & \underline{89.5} & \textbf{79.7} & \underline{80.5} \\
\cmidrule(lr){2-6}
& GenAU-0.5b &   92.0 & 84.5 & 66.0 & 73.5 \\
& GenAU-7b &       94.5 & 87.1 & 70.3 & 78.5 \\
\midrule
\multirow{8}{*}{Real-IAD} 
& WinCLIP      & 87.1 & 59.9 & 75.0 & 72.3 \\
& APRIL-GAN    & 96.0 & 86.8 & 75.7 & 73.5 \\
& AnomalyCLIP  & 96.2 & 85.7 & 78.4 & 76.7 \\
& AdaCLIP      & 95.3 & - & 70.1 & 68.5 \\
& MultiADS   & \textbf{96.6} & \underline{87.1} & \underline{78.7} & \underline{79.1 }\\
& MultiADS-F   &  \underline{96.3} & \textbf{87.2} & 78.2 & 78.5 \\
\cmidrule(lr){2-6}
& GenAU-0.5b &        87.4 & 73.3 & 76.6 & 78.1\\
& GenAU-7b &        94.8 & \underline{87.1} & \textbf{84.7} & \textbf{85.2}\\
\bottomrule
\end{tabular}
\end{table}
Qualitative results of GenAU and the best overall performer, MultiADS, are shown in Figure~\ref{fig:visa-visual-good} (and, for a failure case, Figure~\ref{fig:visa-visual-bad}). GenAU accurately segments visually salient defects such as \textit{hole}, \textit{broken}, and \textit{crack}, and produces cleaner, lower-noise segmentation maps than MultiADS. However, global-context or logic-dependent defects such as \textit{missing} components remain challenging. For these cases, similarity maps are low-confidence, often resulting in incomplete localization or misclassification of normal regions for both models (Figure~\ref{fig:visa-visual-bad}).

\textbf{Ablation Studies}.
Table~\ref{tab:ablations_main} varies one factor at a time from our default GenAU-7B on VisA. Removing the LM fusion (\emph{map-max only}) lowers image-AUROC by $2.3$ points ($87.6\!\rightarrow\!85.3$); the segmentation-map maximum alone is already a competitive detector (above the CLIP baselines), with the decoder log-odds adding a consistent refinement rather than carrying the score. Adding the \emph{reasoning} instruction mix to training (the setting also used for defect reasoning in Table~\ref{tab:reasoning}) costs only $2.9$ image-AUROC and $2.6$ pixel-AUPRO, so the model retains most of its detection and localization while gaining reasoning, the segmentation--reasoning trade-off. Dropping LoRA on the LLM costs a further $2.2$ image-AUROC and $1.1$ pixel-AUPRO, confirming its contribution to detection and localization. The ablation studies of hyper-parameters are in Appendix.

\begin{table}
\small
\centering
\setlength{\tabcolsep}{3pt}
\caption{GenAU-7B ablations on VisA. The top (shaded) row is our default; each subsequent row changes a single factor relative to it. Fusion affects only the image score, so its pixel columns are unchanged. All metrics in \%.}
\label{tab:ablations_main}
\begin{tabular}{@{}l|cc|cc@{}}
\toprule
 & \multicolumn{2}{c|}{\textbf{Pixel}} & \multicolumn{2}{c}{\textbf{Image}} \\
\cmidrule(lr){2-3}\cmidrule(lr){4-5}
 & AUROC & AUPRO & AUROC & AP \\
\midrule
\rowcolor{gray!20}
GenAU-7B & 94.7 & 88.2 & 87.6 & 89.9 \\
\midrule
\quad w/o LM fusion & 94.7 & 88.2 & 85.3 & 87.6 \\
\quad w/ reasoning mix & 95.1 & 85.6 & 84.7 & 87.7 \\
\quad w/o LoRA & 94.8 & 87.1 & 85.4 & 87.1 \\
\bottomrule
\end{tabular}
\end{table}

\subsubsection{Zero-Shot Multi-Type Detection}
We evaluate zero-shot multi-type defect classification on VisA and MPDD (Table~\ref{tab:multytype_zero_shot_detection_performance}), conditioning the model on the target object's candidate defect list (a candidate-list-conditioned zero-shot setting in Appendix). F1 ranges from $\sim$$29\%$ on VisA to $\sim$$70\%$ on MPDD, the latter higher owing to its fewer defect types. Most prior methods target only binary detection, limiting comparison; MultiADS is the only multi-type-capable baseline.
GenAU-7B achieves the best F1 on both datasets, outperforming MultiADS and the reasoning specialist Anomaly-OV at both scales (Anomaly-OV-0.5b/7b are our MVTec-trained reproductions, prompted in a two-turn detect-then-classify format), despite spending no training budget on reasoning; Anomaly-OV is comparatively strong on MPDD (few types, where Anomaly-OV-7B even leads on recall) but weaker on VisA, whose finer taxonomy is harder. One caveat tempers the MultiADS comparison: MultiADS, designed for multi-type \emph{segmentation}, is converted to detection by thresholding its per-type channels at a single fixed value of $0.5$, an operating point that may understate it. GenAU's advantage is driven mainly by precision (e.g., $50.3$ vs MultiADS's $17.4$ on VisA and $90.4$ vs $34.0$ on MPDD), reflecting fewer spurious defect predictions under candidate-list-constrained decoding. Absolute scores remain low partly because VisA has 14 defect types and, with multi-defect images, 56 unique combinations to predict.

\begin{table}
\small
\centering
\caption[Multi-type zero-shot detection performance of GenAU-7b.]{Multi-type zero-shot detection performance on VisA and MPDD. All metrics are in \%. \textbf{Bold} marks the best performer per column (within each dataset). Anomaly-OV-0.5b and -7b are our MVTec-trained reproductions, prompted in a two-turn detect-then-classify format suited to their conversational training.}
\label{tab:multytype_zero_shot_detection_performance}
\begin{tabular}{@{}llccc@{}}
\toprule
\textbf{Dataset}          & \textbf{Model} & \textbf{Precision} & \textbf{Recall} & \textbf{F1} \\ \midrule
\multirow{5}{*}{VisA}     & MultiADS         & 17.4          & \textbf{57.7} & 18.6          \\
                          & Anomaly-OV-0.5b  & 18.4          & 11.1          & 9.3           \\
                          & Anomaly-OV-7b  & 42.1          & 30.9          & 21.1           \\
                          \cline{2-5}
                          & GenAU-0.5b         & 14.8 & 17.2          & 11.1 \\
                          & GenAU-7b         & \textbf{50.3} & 34.1          & \textbf{29.0} \\ \midrule
\multirow{5}{*}{MPDD}     & MultiADS         & 34.0          & 51.3          & 32.7          \\
                          & Anomaly-OV-0.5b  & 60.1          & 30.5           & 29.9          \\
                          & Anomaly-OV-7b  & 87.1          & \textbf{65.8}           & 65.1          \\
                          \cline{2-5}
                          & GenAU-0.5b         & 65.2 & 40.4          & 38.8 \\
                          & GenAU-7b         & \textbf{90.4} & 64.2 & \textbf{69.8} \\
\bottomrule
\end{tabular}
\end{table}

\subsubsection{Defect Reasoning and Analysis}
Beyond detecting and localizing defects, GenAU answers open-ended questions about them. We evaluate on VisA-D\&R~\cite{xu2025towards} (low-level perception and complex causal-reasoning questions) against the reasoning specialist Anomaly-OV~\cite{xu2025towards}; here GenAU augments its segmentation training with a reasoning instruction mix, so one model both localizes and explains, whereas Anomaly-OV produces no segmentation.

As shown in Table~\ref{tab:reasoning}, GenAU improves markedly over its own LLaVA-OneVision backbone (at 7B, low-level perception rises from $3.57$ to $5.22$ and complex reasoning from $5.44$ to $6.43$) and exceeds the dedicated specialist Anomaly-OV at both scales. Strikingly, even GenAU-0.5B ($5.01$/$6.13$) surpasses the 7B specialist ($4.26$/$6.34$) on low-level perception, showing the reasoning instruction mix is effective even at small scale. This indicates that augmenting a segmentation-capable VLM with a lightweight reasoning instruction mix yields strong conversational ability without training a dedicated reasoning model, while still emitting the segmentation tokens needed for localization. These baseline scores, including our LLaVA-OneVision backbone, are taken from~\cite{xu2025towards} under a separate GPT-4o judge instance, so the absolute comparison is indicative rather than strictly controlled; the consistent margin across both question types and both scales is what we rely on. More importantly, once trained with the reasoning instruction mix, GenAU is the only model in our study that performs binary detection, pixel-level segmentation, multi-type recognition, \emph{and} defect reasoning (Table~\ref{tab:ablations_main}): the CLIP baselines emit no language and the reasoning specialist no masks. This supports our central claim that one \emph{architecture and training recipe}, with an explicit segmentation--reasoning trade-off, covers the full inspection stack rather than a bespoke model per task.

\begin{table}
\small
\centering
\setlength{\tabcolsep}{6pt}
\caption{Defect reasoning on VisA-D\&R~\cite{xu2025towards}, scored by GPT-4o on a $1$--$10$ scale. VisA-D\&R poses two question categories per image: \emph{Low-level} perception questions, which ask the model to describe the anomaly, and \emph{Complex} causal-reasoning questions about its significance, likely cause, and prevention; we report the GPT-Score separately for each. $^\dagger$LLaVA-OV and Anomaly-OV scores as reported in the Anomaly-OV paper~\cite{xu2025towards}. For this task GenAU augments its segmentation training with a reasoning instruction mix, and is evaluated as a single model that both localizes and explains anomalies.}
\label{tab:reasoning}
\begin{tabular}{@{}lcc@{}}
\toprule
\multirow{2}{*}{\textbf{Model}} & \multicolumn{2}{c}{\textbf{GPT-Score}} \\
\cmidrule(lr){2-3}
 & \textbf{Low-level} & \textbf{Complex} \\
\midrule
LLaVA-OV-0.5B$^\dagger$   & 2.54 & 4.34 \\
LLaVA-OV-7B$^\dagger$     & 3.57 & 5.44 \\
Anomaly-OV-0.5B$^\dagger$ & 3.87 & 5.67 \\
Anomaly-OV-7B$^\dagger$   & 4.26 & 6.34 \\
\cmidrule(lr){1-3}
GenAU-0.5B & 5.01          & 6.13          \\
GenAU-7B   & \textbf{5.22} & \textbf{6.43} \\
\bottomrule
\end{tabular}
\end{table}

\section{Discussion}
This section provides the performance analyses of GenAU, highlighting challenges and insights that are often unexplored in prior studies.

\subsection{Local vs. Global Context Segmentation}
\begin{figure}[t]
  \centering
  \includegraphics[width=1.0\columnwidth]{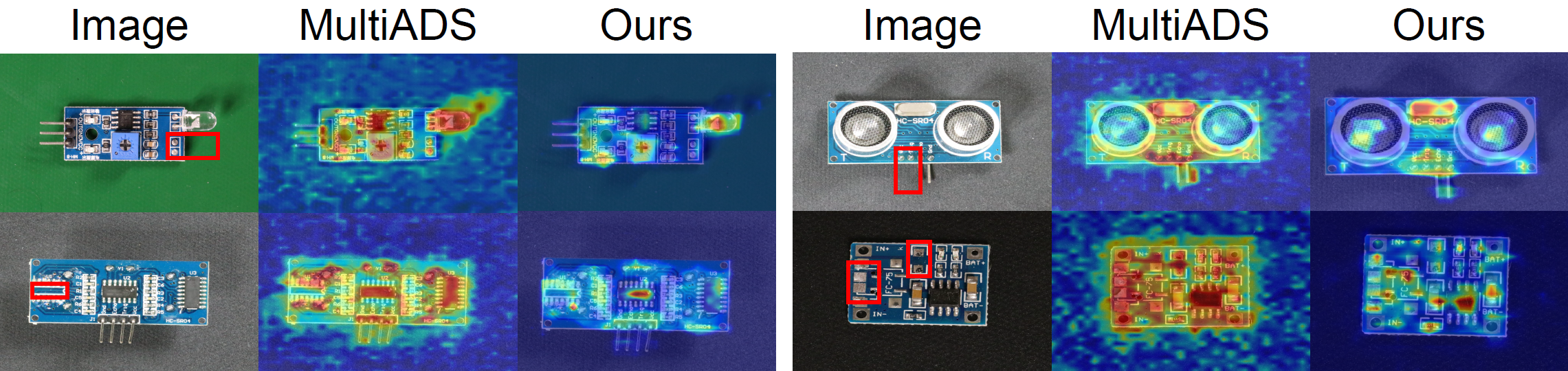}
  \caption{Anomaly segmentation on VisA for the \emph{missing} defect, a globally-defined anomaly, with MultiADS for comparison. Both methods yield low-confidence, incomplete maps.}
  \label{fig:visa-visual-bad}
\end{figure}
GenAU excels at visually salient anomalies (distinct texture or color deviations), where strong local cues align with the defect-token anchors. Segmentation degrades for globally-defined or logical anomalies (e.g., \textit{missing} or \textit{misplaced} components; Figure~\ref{fig:visa-visual-bad}), which look normal at the patch level and, in the zero-shot setting, give no prior on expected placement; few-shot or domain priors could help.

\subsection{Challenges in Defect-Type Classification}
Fine-grained classification exposes the limits of the VLM's pre-trained knowledge: the same label varies widely across objects (a \textit{missing cable} is a conspicuous void, a \textit{missing LED} blends in; cracks range from large openings to fine lines), so generalization is hard when training shows one instance of a defect. Logical defects further require reasoning about expected structure that the model cannot encode without domain knowledge; broader exposure to each defect would help.

\section{Conclusion}
We presented GenAU, a single instruction-following architecture and training recipe that unifies the four questions of industrial inspection (is there an anomaly, where, what type, and why). Across VisA, MPDD, and Real-IAD it achieves competitive zero-shot detection and segmentation under distribution shift and, when the same recipe additionally learns to reason, adds multi-type recognition and defect analysis at a small, quantified cost. Globally-defined anomalies and fine-grained, object-dependent defect types remain open, pointing to few-shot adaptation and richer domain priors.

\section*{Acknowledgments}
The authors thank the International Max Planck Research School for Intelligent Systems (IMPRS-IS)
for supporting Hongkuan Zhou and Jingcheng Wu.

{
    \small
    \bibliographystyle{ieeenat_fullname}
    \bibliography{custom}
}

\appendix

\section{Datasets}
\label{app:datasets}
Due to space limitations in the main manuscript, we provide here a detailed description of the industrial anomaly detection datasets used in our experiments: MVTec-AD~\cite{Bergmann_2019_CVPR}, VisA~\cite{VisA}, MPDD~\cite{mpdd}, and Real-IAD~\cite{wang2024realiadrealworldmultiviewdataset}. Key dataset statistics, including the number of defect types, object categories, and the proportion of single- and multi-defect images, are summarized in \cref{tab:dataset_summary}. For all datasets, each defect instance is mapped to a unified defect taxonomy as defined in \crefrange{tab:mvtec-mapping1}{tab:real-iad-mapping2}.
MVTec-AD includes common industrial items such as bottles, cables, zippers, and hazelnuts. VisA contains a mixture of everyday products such as chewing gum and macaroni, as well as more complex components like printed circuit boards. MPDD focuses primarily on industrial parts, including metal brackets, tubes, and machined components. Real-IAD contains a broad range of real-world consumer and industrial items such as audio jacks, switches, bottle caps, and plastic toys.

\begin{table*}
\centering
\setlength{\tabcolsep}{5pt}
\caption[Overview of the datasets used for evaluation.]{Overview of the datasets used for evaluation. The third column lists distinct defects, while the fourth column counts compound defects treated as separate categories.}
\label{tab:dataset_summary}
\begin{tabular}{cccccc}
\toprule
\textbf{Dataset} & \textbf{Objects} & \textbf{\begin{tabular}[c]{@{}c@{}}Unique \\ Defect Types\end{tabular}} & \textbf{\begin{tabular}[c]{@{}c@{}}Defect Types \\ Incl. Combinations\end{tabular}} & \textbf{\begin{tabular}[c]{@{}c@{}}Single Defect \\ Images\end{tabular}} & \textbf{\begin{tabular}[c]{@{}c@{}}Multi Defect \\ Images\end{tabular}} \\
\midrule
MVTec-AD  & 15 & 16 & 38   & 1,200  & 58   \\
VisA      & 12 & 14 & 56   & 923    & 277  \\
MPDD      & 6  & 6  & 7    & 265    & 17   \\
Real-IAD  & 30 & 9  & 9    & 51,329 & 0    \\
\bottomrule
\end{tabular}
\end{table*}

\begin{table}
\centering
\footnotesize
\caption{Defect taxonomy for the MVTec-AD dataset (Part I).}
\label{tab:mvtec-mapping1}
\begin{tabular}{@{}ccc@{}}
\toprule
\textbf{Product} & \textbf{Original Defect Types} & \textbf{Mapped Defect Types} \\
\midrule

\multirow{3}{*}{Bottle}
 & Broken Large & Broken \\
 & Broken Small & Broken \\
 & Contamination & Contamination \\

\midrule
\multirow{8}{*}{Cable}
 & Bent Wire & Bent \\
 & Cable Swap & Misplaced \\
 & Combined & Combination of Defects \\
 & Cut Inner Insulation & Cut \\
 & Cut Outer Insulation & Cut \\
 & Missing Cable & Missing \\
 & Missing Wire & Missing \\
 & Poke Insulation & Poke \\

\midrule
\multirow{5}{*}{Capsule}
 & Crack & Crack \\
 & Faulty Imprint & Scratch \\
 & Poke & Crack \\
 & Scratch & Scratch \\
 & Squeeze & Deformed \\

\midrule
\multirow{5}{*}{Carpet}
 & Color & Color \\
 & Cut & Tear \\
 & Hole & Hole \\
 & Metal Contamination & Contamination \\
 & Thread & Contamination \\

\midrule
\multirow{4}{*}{Hazelnut}
 & Crack & Crack \\
 & Cut & Cut \\
 & Hole & Hole \\
 & Print & Color \\

\midrule
\multirow{5}{*}{Grid}
 & Bent & Bent \\
 & Broken & Broken \\
 & Glue & Contamination \\
 & Metal Contamination & Contamination \\
 & Thread & Contamination \\

\midrule
\multirow{5}{*}{Leather}
 & Color & Color \\
 & Cut & Cut \\
 & Fold & Fold \\
 & Glue & Contamination \\
 & Poke & Poke \\

\midrule
\multirow{4}{*}{Metal Nut}
 & Bent & Bent \\
 & Color & Color \\
 & Flip & Misplaced \\
 & Scratch & Scratch \\

\midrule
\multirow{7}{*}{Pill}
 & Color & Color \\
 & Combined & Combination of Defects \\
 & Contamination & Contamination \\
 & Crack & Crack \\
 & Faulty Imprint & Scratch \\
 & Pill Type & Color \\
 & Scratch & Scratch \\

\midrule
\multirow{5}{*}{Screw}
 & Manipulated Front & Bent Head / Broken Tip \\
 & Scratch Head & Damaged \\
 & Scratch Neck & Damaged \\
 & Thread Side & Damaged \\
 & Thread Top & Damaged \\

\bottomrule
\end{tabular}
\end{table}

\begin{table}
\centering
\footnotesize
\caption{Defect taxonomy for the MVTec-AD dataset (Part II).}
\label{tab:mvtec-mapping2}
\begin{tabular}{@{}ccc@{}}
\toprule
\textbf{Product} & \textbf{Original Defect Types} & \textbf{Mapped Defect Types} \\
\midrule

\multirow{5}{*}{Tile}
 & Crack & Crack \\
 & Glue Strip & Contamination \\
 & Gray Stroke & Color \\
 & Oil & Contamination \\
 & Rough & Scratch \\

\midrule
\multirow{1}{*}{Toothbrush}
 & Defective & Bent / Contamination / Missing \\

\midrule
\multirow{4}{*}{Transistor}
 & Bent Lead & Bent \\
 & Cut Lead & Broken \\
 & Damaged Case & Damaged \\
 & Misplaced & Misplaced \\

\midrule
\multirow{5}{*}{Wood}
 & Color & Color \\
 & Combined & Combination of Defects \\
 & Hole & Hole \\
 & Liquid & Contamination \\
 & Scratch & Scratch \\

\midrule
\multirow{7}{*}{Zipper}
 & Broken Teeth & Broken \\
 & Combined & Combination of Defects \\
 & Fabric Border & Tear \\
 & Fabric Interior & Tear \\
 & Rough & Contamination \\
 & Split Teeth & Deformed \\
 & Squeezed Teeth & Deformed \\

\bottomrule
\end{tabular}
\end{table}

\begin{table}
\centering
\scriptsize
\renewcommand{\arraystretch}{0.9}
\caption{Defect taxonomy for the VisA dataset (Part I).}
\label{tab:visa-mapping1}
\begin{tabular}{@{}ccc@{}}
\toprule
\textbf{Product} & \textbf{Original Defect Types} & \textbf{Mapped Defect Types} \\
\midrule

\multirow{8}{*}{Candle}
 & Chunk of Wax Missing & Hole \\
 & Damaged Corner of Packaging & Damaged \\
 & Different Colour Spot & Color \\
 & Extra Wax in the Candle & Contamination \\
 & Foreign Particals on the Candle & Color \\
 & Wax Melded out of the Candle & Melded \\
 & Weird Candle Wick & Weird Wick \\
 & Combined & Combination of Defects \\

\midrule
\multirow{6}{*}{Capsules}
 & Bubble & Bubble \\
 & Discolor & Color \\
 & Scratch & Scratch \\
 & Leak & Contamination \\
 & Misshape & Deformed \\
 & Combined & Combination of Defects \\

\midrule
\multirow{9}{*}{Cashew}
 & Burnt & Color \\
 & Corner or Edge Breakage & Broken \\
 & Different Colour Spot & Color \\
 & Middle Breakage & Broken \\
 & Same Colour Spot & Color \\
 & Small Holes & Hole \\
 & Small Scratches & Scratch \\
 & Stuck Together & Misplaced \\
 & Combined & Combination of Defects \\

\midrule
\multirow{6}{*}{Chewinggum}
 & Chunk of Gum Missing & Broken \\
 & Corner Missing & Broken \\
 & Scratches & Scratch \\
 & Similar Colour Spot & Color \\
 & Small Cracks & Crack \\
 & Combined & Combination of Defects \\

\midrule
\multirow{8}{*}{Fryum}
 & Burnt & Color \\
 & Corner or Edge Breakage & Broken \\
 & Different Colour Spot & Color \\
 & Fryum Stuck Together & Misplaced \\
 & Middle Breakage & Broken \\
 & Similar Colour Spot & Color \\
 & Small Scratches & Scratch \\
 & Combined & Combination of Defects \\

\bottomrule
\end{tabular}
\end{table}

\begin{table}
\centering
\scriptsize
\renewcommand{\arraystretch}{0.9}
\caption{Defect taxonomy for the VisA dataset (Part II).}
\label{tab:visa-mapping2}
\begin{tabular}{@{}ccc@{}}
\toprule
\textbf{Product} & \textbf{Original Defect Types} & \textbf{Mapped Defect Types} \\
\midrule

\multirow{7}{*}{Macaroni1}
 & Chip Around Edge and Corner & Broken \\
 & Different Colour Spot & Color \\
 & Similar Colour Spot & Color \\
 & Small Cracks & Crack \\
 & Middle Breakage & Hole \\
 & Small Scratches & Scratch \\
 & Combined & Combination of Defects \\

\midrule

\multirow{7}{*}{Macaroni2}
 & Breakage down the Middle & Hole \\
 & Color Spot Similar to the Object & Color \\
 & Different Color Spot & Color \\
 & Small Chip Around Edge & Broken \\
 & Small Cracks & Crack \\
 & Small Scratches & Scratches \\
 & Combined & Combination of Defects \\

\midrule
\multirow{5}{*}{pcb1}
 & Bent & Bent \\
 & Melt & Contamination \\
 & Missing & Missing \\
 & Scratch & Scratch \\
 & Combined & Combination of Defects \\

\midrule
\multirow{5}{*}{pcb2}
 & Bent & Bent \\
 & Melt & Contamination \\
 & Missing & Missing \\
 & Scratch & Scratch \\
 & Combined & Combination of Defects \\

\midrule
\multirow{5}{*}{pcb3}
 & Bent & Bent \\
 & Melt & Contamination \\
 & Missing & Missing \\
 & Scratch & Scratch \\
 & Combined & Combination of Defects \\

\midrule
\multirow{8}{*}{pcb4}
 & Burnt & Contamination \\
 & Scratch & Scratch \\
 & Damage & Damaged \\
 & Dirt & Contamination \\
 & Extra & Contamination \\
 & Missing & Missing \\
 & Wrong Place & Contamination \\
 & Combined & Combination of Defects \\

\midrule
\multirow{8}{*}{Pipe Fryum}
 & Burnt & Color \\
 & Corner and Edge Breakage & Broken \\
 & Different Colour Spot & Color \\
 & Similar Colour Spot & Color \\
 & Small Scratches & Scratch \\
 & Stuck Together & Misplaced \\
 & Small Cracks & Crack \\
 & Combined & Combination of Defects \\

\bottomrule
\end{tabular}
\end{table}

\begin{table}
\centering
\caption{Defect taxonomy for the MPDD dataset.}
\footnotesize
\label{tab:mpdd-mapping}
\begin{tabular}{@{}ccc@{}}
\toprule
\textbf{Product} & \textbf{Original Defects Types}                                                 & \textbf{Mapped Defect Types}                                           \\ \midrule
Bracket Black    & \begin{tabular}[c]{@{}c@{}}Hole\\ Scratch\end{tabular}                           & \begin{tabular}[c]{@{}c@{}}Hole\\ Scratch\end{tabular}                 \\ \midrule
Bracket Brown    & \begin{tabular}[c]{@{}c@{}}Bend and Parts Mismatch\\ Parts Mismatch\end{tabular} & \begin{tabular}[c]{@{}c@{}}Bent and Misplaced\\ Misplaced\end{tabular} \\ \midrule
Bracket White    & \begin{tabular}[c]{@{}c@{}}Defective Painting\\ Scratches\end{tabular}           & \begin{tabular}[c]{@{}c@{}}Contamination\\ Scratch\end{tabular}        \\ \midrule
Connector        & Parts Mismatch                                                                   & Missing                                                                \\ \midrule
Metal Plate      & \begin{tabular}[c]{@{}c@{}}Major Rust\\ Total Rust\\ Scratch\end{tabular}        & \begin{tabular}[c]{@{}c@{}}Color\\ Color\\ Scratch\end{tabular}        \\ \midrule
Tubes            & Anomalous                                                                        & Deformed                                                               \\ \bottomrule
\end{tabular}
\end{table}

\begin{table}
\centering
\footnotesize
\renewcommand{\arraystretch}{0.7}
\caption{Defect taxonomy for the Real-IAD dataset (Part I).}
\label{tab:real-iad-mapping1}
\begin{tabular}{@{}ccc@{}}
\toprule
\textbf{Product} & \textbf{Original Defect Types} & \textbf{Mapped Defect Types} \\
\midrule

\multirow{4}{*}{Audiojack}
 & Contamination & Color \\
 & Missing Parts & Broken \\
 & Deformation & Bent \\
 & Scratch & Scratch \\

\midrule
\multirow{4}{*}{Bottle Cap}
 & Contamination & Color \\
 & Missing Parts & Broken \\
 & Scratch & Scratch \\
 & Pit & Hole \\

\midrule
\multirow{4}{*}{Button Battery}
 & Scratch & Scratch \\
 & Abrasion & Abrasion \\
 & Contamination & Color \\
 & Pit & Color \\

\midrule
\multirow{4}{*}{End Cap}
 & Damage & Broken \\
 & Missing Parts & Broken \\
 & Contamination & Color \\
 & Scratch & Scratch \\

\midrule
\multirow{4}{*}{Eraser}
 & Contamination & Color \\
 & Scratch & Scratch \\
 & Pit & Hole \\
 & Missing Parts & Broken \\

\midrule
\multirow{4}{*}{Fire Hood}
 & Pit & Hole \\
 & Contamination & Color \\
 & Missing Parts & Broken \\
 & Scratch & Scratch \\

\midrule
\multirow{3}{*}{Mint}
 & Missing Parts & Broken \\
 & Foreign Objects & Contamination \\
 & Contamination & Contamination \\

\midrule
\multirow{3}{*}{Mounts}
 & Missing Parts & Broken \\
 & Contamination & Color \\
 & Pit & Hole \\

\midrule
\multirow{4}{*}{PCB}
 & Missing Parts & Missing \\
 & Scratch & Scratch \\
 & Foreign Objects & Contamination \\
 & Contamination & Contamination \\

\midrule
\multirow{4}{*}{Phone Battery}
 & Damage & Broken \\
 & Pit & Hole \\
 & Scratch & Scratch \\
 & Contamination & Contamination \\

\midrule
\multirow{4}{*}{Plastic Nut}
 & Pit & Hole \\
 & Scratch & Scratch \\
 & Contamination & Contamination \\
 & Damage & Broken \\

\midrule
\multirow{4}{*}{Plastic Plug}
 & Missing Parts & Broken \\
 & Scratch & Scratch \\
 & Pit & Hole \\
 & Contamination & Contamination \\

\midrule
\multirow{3}{*}{Porcelain Doll}
 & Scratch & Scratch \\
 & Abrasion & Abrasion \\
 & Contamination & Color \\

\bottomrule
\end{tabular}
\end{table}

\begin{table}
\centering
\footnotesize
\renewcommand{\arraystretch}{0.7}
\caption{Defect taxonomy for the Real-IAD dataset (Part II).}
\label{tab:real-iad-mapping2}
\begin{tabular}{@{}ccc@{}}
\toprule
\textbf{Product} & \textbf{Original Defect Types} & \textbf{Mapped Defect Types} \\
\midrule

\multirow{4}{*}{Regulator}
 & Contamination & Color \\
 & Missing Parts & Broken \\
 & Scratch & Scratch \\
 & Pit & Hole \\

\midrule
\multirow{3}{*}{Rolled Strip Base}
 & Pit & Hole \\
 & Missing Parts & Broken \\
 & Contamination & Color \\

\midrule
\multirow{3}{*}{SIM Card Set}
 & Scratch & Scratch \\
 & Abrasion & Scratch \\
 & Contamination & Contamination \\

\midrule
\multirow{3}{*}{Switch}
 & Missing Parts & Broken \\
 & Contamination & Contamination \\
 & Scratch & Scratch \\

\midrule
\multirow{4}{*}{Tape}
 & Scratch & Scratch \\
 & Missing Parts & Broken \\
 & Damage & Broken \\
 & Contamination & Color \\

\midrule
\multirow{3}{*}{Terminalblock}
 & Pit & Hole \\
 & Missing Parts & Broken \\
 & Contamination & Color \\

\midrule
\multirow{3}{*}{Toothbrush}
 & Missing Parts & Broken \\
 & Contamination & Contamination \\
 & Abrasion & Deformed \\

\midrule
\multirow{4}{*}{Toy}
 & Scratch & Scratch \\
 & Contamination & Color \\
 & Missing Parts & Broken \\
 & Pit & Hole \\

\midrule
\multirow{4}{*}{Toy Brick}
 & Contamination & Color \\
 & Pit & Hole \\
 & Missing Parts & Broken \\
 & Scratch & Scratch \\

\midrule
\multirow{3}{*}{Transistor1}
 & Missing Parts & Broken \\
 & Deformation & Bent \\
 & Contamination & Contamination \\

\midrule
\multirow{4}{*}{U Block}
 & Missing Parts & Broken \\
 & Scratch & Scratch \\
 & Abrasion & Scratch \\
 & Contamination & Color \\

\midrule
\multirow{4}{*}{USB}
 & Scratch & Scratch \\
 & Deformation & Bent \\
 & Contamination & Color \\
 & Missing Parts & Broken \\

\midrule
\multirow{4}{*}{USB Adaptor}
 & Contamination & Color \\
 & Scratch & Scratch \\
 & Abrasion & Scratch \\
 & Pit & Broken \\

\midrule
\multirow{4}{*}{Vcpill}
 & Contamination & Color \\
 & Scratch & Scratch \\
 & Missing Parts & Broken \\
 & Pit & Broken \\

\midrule
\multirow{4}{*}{Wooden Beads}
 & Scratch & Scratch \\
 & Contamination & Color \\
 & Pit & Hole \\
 & Missing Parts & Broken \\

\midrule
\multirow{4}{*}{Woodstick}
 & Contamination & Color \\
 & Scratch & Scratch \\
 & Pit & Hole \\
 & Missing Parts & Broken \\

\midrule
\multirow{4}{*}{Zipper}
 & Missing Parts & Broken \\
 & Damage & Broken \\
 & Deformation & Deformed \\
 & Contamination & Color \\

\bottomrule
\end{tabular}
\end{table}

\section{Implementation Details}
\label{app:impl}
We build GenAU on the LLaVA-OneVision family~\cite{li2025llavaonevision}, which couples a Qwen2 language backbone with a SigLIP-so400m-patch14-384 vision encoder, and report the 0.5B and 7B scales. High-resolution inputs use the AnyRes scheme (\texttt{anyres\_max\_9}) with an image grid of up to $6\times6$ tiles of $384$ pixels, in addition to a global low-resolution view. For the segmentation readout we take patch features from four vision-encoder stages (layers $6$, $12$, $18$, and the final layer) and equip each stage with its own linear projection into the decoder hidden space.

\paragraph{Trainable parameters.}
The vision encoder and the original cross-modal projector are frozen. The LLM is adapted with rank-stabilized LoRA (rank $r=16$, $\alpha=32$, dropout $0$). The token-embedding table and language-modeling head are fully fine-tuned to integrate the two added segmentation tokens \texttt{[SEG\_normal]} and \texttt{[SEG\_defect]}, and the four segmentation projectors together with the decoder layer-norms are trainable.

\paragraph{Optimization.}
Training runs for a single epoch with an effective batch size of $32$ (per-device batch $1$, gradient accumulation $4$, across $8$ GPUs with DeepSpeed ZeRO-3), a learning rate of $10^{-4}$ with cosine decay, a warmup ratio of $0.03$, and no weight decay. The segmentation objective sums, over the four stages, a focal term on the upsampled logits, a Tversky term ($\alpha=0.3$, $\beta=0.7$) on the defect channel that up-weights false negatives, and a Dice term on the complementary normal channel; the text-to-segmentation weight is $\lambda=1$.

\paragraph{High-resolution readout and image-level fusion.}
At inference the high-resolution and global-view anomaly maps are fused per stage as $\mathbf{A}=(1-\beta)\,\mathbf{A}^{\mathrm{g}}+\beta\,\mathbf{A}^{\mathrm{hi}}$ with $\beta=0.25$ and a similarity temperature of $20$. The image-level score fuses the spatial maximum of $\mathbf{A}$ with the decoder's normal/defect log-odds: the two are each $z$-score standardized with statistics estimated once on the MVTec calibration split and summed with a mixing weight of $0.5$; the standardization is frozen and shared across target datasets, so no target-domain labels are used.

\section{Training Data Construction}
\label{app:data}
Training data are derived from established anomaly detection and segmentation benchmarks~\cite{Bergmann_2019_CVPR, VisA, wang2024realiadrealworldmultiviewdataset}. We retain the original images and segmentation masks and augment them with instruction--response pairs designed for instruction tuning; this augmentation is the only modification to the original data. Each sample thus consists of the original image, its segmentation mask, a textual instruction, and the corresponding response. The instruction asks, via natural-language paraphrases, whether the named object contains an anomaly and requests its localization, e.g.\ \texttt{``Is there any anomaly on the \{object\}? If yes, segment it.''}

The response describes the anomaly in natural language and terminates with the two segmentation tokens \texttt{[SEG\_defect]} and \texttt{[SEG\_normal]}, from which the localization read-out is taken; for a normal image it instead states that no anomaly is present. An example response is:

\textit{``Yes, there is an anomaly on the capsule. Specifically, there is a dent on the orange side, affecting the clarity and smoothness of the numeral `0' in the printed number `500.' [SEG\_defect] [SEG\_normal]''}

\section{Defect Taxonomy}
\label{app:taxonomy}
Industrial AD datasets often define highly specific defect categories, which can lead to overfitting and inconsistent label usage~\cite{mokhtar2025detectclassifyactcategorizing}. To improve generalization, we map product-specific labels to a unified, coarser taxonomy, following MultiADS~\cite{sadikaj2025multiadsdefectawaresupervisionmultitype}: e.g., \textit{liquid} and \textit{metal contamination} are merged into \textit{contamination}, and \textit{broken large}/\textit{broken small} are unified as \textit{broken}. The taxonomy is defined per object, so the candidate defect set differs across product categories. This reclassification promotes abstraction over product-specific anomalies and mitigates label imbalance, and it does not affect comparisons with prior binary-detection methods, whose metrics are independent of class granularity. The complete per-object mapping for all four datasets is given in \crefrange{tab:mvtec-mapping1}{tab:real-iad-mapping2}.

\section{Evaluation Protocol}
\label{app:eval}
All models are evaluated under single-image inference: only the query image is available at test time, with no reference, support, or exemplar images. GenAU is fine-tuned on the MVTec-AD test split (the only MVTec split carrying anomaly masks and defect-type labels, as the train split contains normal images only) and evaluated, without retraining, on the test splits of VisA, MPDD, and Real-IAD; MVTec thus serves purely as the source domain and is disjoint from all evaluation targets, so no test-time information leaks into the reported numbers. For Real-IAD we use the single-view setting.

\paragraph{Unified inspection prompt.}
For binary detection and segmentation, the instruction names the object category and asks whether the image contains an anomaly, requesting its segmentation. The model generates a textual answer immediately followed by the two segmentation tokens, e.g.\ \texttt{\{defect\} [SEG\_defect] [SEG\_normal]} for an anomalous image and a statement of normality followed by the same tokens for a normal one; both \texttt{[SEG\_defect]} and \texttt{[SEG\_normal]} are \emph{always} emitted, and we extract their final-layer hidden states as language-grounded anchors.

\paragraph{Product-conditioned multi-type detection.}
For zero-shot multi-type detection, the instruction is conditioned on the product type: it lists the candidate defect types of the \emph{target object} (its coarse classes under the taxonomy above) and asks the model to select one. A deterministic parser maps the generated response to a multi-type label over the target-domain defect vocabulary. We report weighted Precision, Recall, and F1 to account for the strong class imbalance across defect categories.

\section{Additional Ablations}
\label{app:ablation}
Table~\ref{tab:ablations} ablates, on MVTec-AD$\rightarrow$VisA, four training choices relative to our default GenAU-7B: the text-to-segmentation loss weight $\lambda$, the number of intermediate vision-encoder stages $m$ supplying the segmentation readout, whether the cross-modal projection to the LLM is trainable, and whether the vision encoder itself is unfrozen. Each row changes a single factor; the shaded default row corresponds to the GenAU-7B configuration of Table~\ref{tab:binary}.

Raising the text-to-segmentation weight to $\lambda=5$ trades a small pixel-AUROC gain for lower image-level detection (image-AUROC $87.6\rightarrow87.0$), so we keep the balanced $\lambda=1$. Reducing the segmentation readout to $m=3$ intermediate stages costs $4.1$ points of pixel-AUPRO ($88.2\rightarrow84.1$) and about a point of image-AUROC, confirming that the multi-stage read-out is load-bearing for localization. Making the cross-modal projection trainable is close to neutral on the pixel metrics but lowers image-AUROC by $1.5$ points; with only the MVTec-AD test split available for fine-tuning, the added capacity yields no gain, so we keep the projection frozen. Finally, unfreezing the vision encoder is markedly harmful, collapsing pixel-AUPRO to $80.6$ and image-AUROC to $73.5$ --- far below every other configuration --- as the encoder over-fits the small fine-tuning set and forgets its general visual features; freezing it is therefore essential.

\begin{table}[h]
\centering
\small
\setlength{\tabcolsep}{3pt}
\caption{Ablation of single training factors on MVTec-AD$\rightarrow$VisA. The shaded row is our default GenAU-7B ($m{=}4$, $\lambda{=}1$, projection and vision encoder frozen, matching Table~\ref{tab:binary}); each subsequent row changes one factor. All metrics in \%.}
\label{tab:ablations}
\begin{tabular}{@{}l|cc|cc@{}}
\toprule
 & \multicolumn{2}{c|}{\textbf{Pixel}} & \multicolumn{2}{c}{\textbf{Image}} \\
\cmidrule(lr){2-3}\cmidrule(lr){4-5}
 & AUROC & AUPRO & AUROC & AP \\
\midrule
\rowcolor{gray!20}
GenAU-7B (default) & 94.7 & 88.2 & 87.6 & 89.9 \\ 
\midrule
\quad $\lambda = 5$            & 95.7 & 88.5 & 87.0 & 89.5 \\ 
\quad $m = 3$ stages          & 94.1 & 84.1 & 86.5 & 88.9 \\ 
\quad projection trainable    & 94.9 & 88.4 & 86.1 & 88.3 \\
\quad vision encoder unfrozen & 94.2 & 80.6 & 73.5 & 77.6 \\
\bottomrule
\end{tabular}
\end{table}

\section{Formal Task Definitions}
\label{app:tasks}
We give the full definitions of the three tasks summarized in Section~\ref{sec:prelims}.

\paragraph{Binary anomaly detection under distribution shift.}
Let $\mathcal{D}_{\text{train}}=\{(\boldsymbol{x}_i,y_i)\}_{i=1}^{N_1}$ and $\mathcal{D}_{\text{target}}=\{(\boldsymbol{x}_i,y_i)\}_{i=N_1+1}^{N_1+N_2}$ be two image--label datasets drawn from different input distributions $\mathcal{P}(\mathcal{X}),\mathcal{P}(\mathcal{X}')$: $\forall i\in\{1,\dots,N_1\}$, $\boldsymbol{x}_i\sim\mathcal{P}(\mathcal{X})$, and $\forall i\in\{N_1+1,\dots,N_1+N_2\}$, $\boldsymbol{x}_i\sim\mathcal{P}(\mathcal{X}')$, where $\mathcal{P}(\mathcal{X})\nsim\mathcal{P}(\mathcal{X}')$. Each image $\boldsymbol{x}\in\mathbb{R}^{H\times W}$ is labeled with $y\in\{0,1\}$ indicating anomaly presence. A binary detector $\hat{f}(\cdot)$ is trained solely on $\mathcal{D}_{\mathrm{train}}$; at test time it is optionally adapted to $\hat{f}'(\cdot)$ using auxiliary target-domain metadata such as a product description. For each $\boldsymbol{x}\in\mathcal{D}_{\mathrm{target}}$, the detector classifies it as normal or anomalous.

\paragraph{Binary anomaly segmentation under distribution shift.}
With $\mathcal{D}_{\text{train}}$ and $\mathcal{D}_{\text{target}}$ as above and $\mathcal{P}(\mathcal{X})\nsim\mathcal{P}(\mathcal{X}')$, each image $\boldsymbol{x}\in\mathbb{R}^{H\times W}$ is paired with a binary mask $\boldsymbol{y}\in\{0,1\}^{H\times W}$ indicating the presence of anomalies at each pixel. The segmentation model $\hat{f}(\cdot)$ is trained solely on $\mathcal{D}_{\mathrm{train}}$ and optionally adapted to $\hat{f}'(\cdot)$ with target-domain metadata. For each pixel of $\boldsymbol{x}\in\mathcal{D}_{\mathrm{target}}$, it classifies the pixel as normal or anomalous.

\paragraph{Zero-shot multi-type anomaly detection.}
Let $\mathcal{D}_{\text{train}}=\{(\boldsymbol{x}_i,\boldsymbol{y}_i)\}_{i=1}^{N_1}$ and $\mathcal{D}_{\text{target}}=\{(\boldsymbol{x}_i,\boldsymbol{y}'_i)\}_{i=N_1+1}^{N_1+N_2}$, with inputs from $\mathcal{P}(\mathcal{X})\nsim\mathcal{P}(\mathcal{X}')$. Each training image carries $\boldsymbol{y}\in\{0,1\}^{K_1+1}$, whose entries index one normal class and $K_1$ defect types $\{d_1,\dots,d_{K_1}\}$; each target image carries $\boldsymbol{y}'\in\{0,1\}^{K_2+1}$ over one normal class and $K_2$ defect types. Writing $C_{\mathrm{train}}=\{d_1,\dots,d_{K_1}\}$ and $C_{\mathrm{target}}=\{d'_1,\dots,d'_{K_2}\}$, we have $C_{\mathrm{train}}\cap C_{\mathrm{target}}\neq\emptyset$ and $C_{\mathrm{target}}\setminus C_{\mathrm{train}}\neq\emptyset$: the target dataset mixes defect types previously seen during training with novel types never seen by the model.

\section{Extended Related Work}
\label{app:related}
This expands the related-work overview of the main text.

\paragraph{CLIP-based prompt adaptation for anomaly detection and segmentation.}
Recent advances in industrial anomaly detection/segmentation have largely built upon CLIP due to its strong zero-shot transfer ability, although CLIP was not originally designed for industrial inspection. A major line of work adapts CLIP to industrial AD through normal/abnormal prompts, learnable prompt tokens, or improved visual--text alignment. WinCLIP~\cite{jeong2023winclip} introduces normal and abnormal textual prompts together with window-based feature aggregation for zero-shot anomaly detection. AnomalyCLIP~\cite{zhou2024anomalyclip} learns object-agnostic prompts to capture generic semantics of normality and abnormality, while PromptAD~\cite{Li_2024_WACV} improves state discrimination by adding learnable prefix and suffix tokens. AdaCLIP~\cite{Cao_2024} further combines static and dynamic learnable prompts for adaptive inference across object categories. Beyond prompt learning, CLIP-based methods have been extended to localization and segmentation by improving patch-level alignment or incorporating external modules. APRIL-GAN~\cite{chen2023zero}, AnoCLIP~\cite{deng2024}, and SimCLIP~\cite{simclip} enhance cross-modal alignment for pixel-level anomaly mapping, while CLIP-SAM~\cite{li2024clipsamclipsamcollaboration} and FiLo~\cite{gu2024filozeroshotanomalydetection} combine CLIP with external segmentation or grounding models, and MultiADS~\cite{sadikaj2025multiadsdefectawaresupervisionmultitype} introduces defect-aware textual supervision for multi-type anomaly detection. Other approaches, such as MuSc~\cite{li2024musczeroshotindustrialanomaly}, MAEDAY~\cite{schwartz2024maeday}, and Aota et al.~\cite{aota2023zero}, rely on patch comparison, feature uniformity, or reconstruction difficulty. Despite their effectiveness, these methods largely depend on CLIP-style embedding comparison, external localization modules, or non-generative anomaly scoring, making them less suited for unified defect reasoning and analysis.

\paragraph{Autoregressive VLMs for anomaly detection and reasoning.}
Autoregressive VLMs provide a different paradigm for anomaly understanding by enabling open-ended visual reasoning and instruction-following interaction. Instead of comparing an image embedding with a fixed set of CLIP prompts, they can generate textual descriptions, answer inspection questions, and reason over defect attributes in natural language. AnomalyGPT~\cite{gu2023anomalygptdetectingindustrialanomalies} is an early effort: it introduces an instruction-following VLM for industrial anomaly reasoning and lets users query the model about abnormal regions, but it relies on an external CLIP-based anomaly map combined with the input image and provided to the language model, so the VLM mainly serves as a reasoning interface on top of a separately generated map. More recently, Anomaly-OV~\cite{xu2025towards} specializes multimodal large language models for zero-shot anomaly detection and defect reasoning via Anomaly-Instruct-125k, a large-scale visual instruction-tuning dataset for anomaly-centric understanding; however, it mainly targets image-level detection and explanatory reasoning, while pixel-level segmentation is not explicitly grounded in the autoregressive decoding process.

\section{Detailed Evaluation Metrics}
\label{app:metrics}
We adopt standard metrics for each task. For image-level anomaly detection, we report the Area Under the Receiver Operating Characteristic Curve (AUROC) and Average Precision (AP). For segmentation, we report the pixel-level AUROC and the Area Under the Per-Region Overlap (AUPRO). For multi-type anomaly detection, we use weighted Precision, Recall, and F1-score to account for the strong class imbalance across defect categories. For defect reasoning, we follow the VisA-D\&R protocol~\cite{xu2025towards} and report a GPT-Score in which GPT-4o rates each response on a $1$--$10$ scale, separately for low-level (perception) and complex (causal-reasoning) questions.

\section{Qualitative Results}
\label{app:qualitative}
\begin{figure*}
  \centering
  \includegraphics[width=\textwidth]{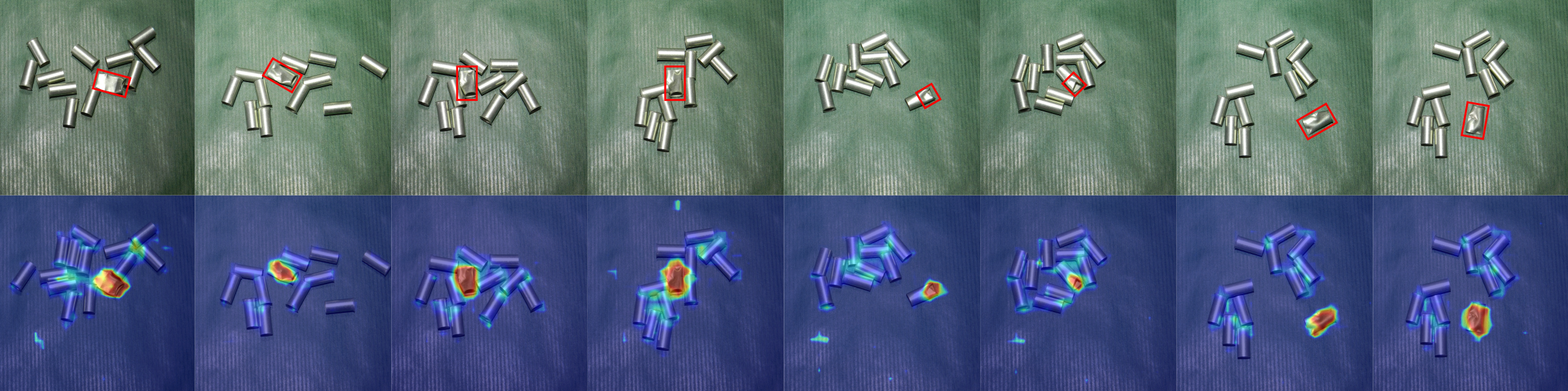}
  \caption{This visualization showcases the tube product from the MPDD dataset. The first row represents the input with the anomaly highlighted. The second row presents the segmentation results
from our model.}
   \label{fig:tubes}
\end{figure*}

\begin{figure*}
  \centering
  \includegraphics[width=\textwidth]{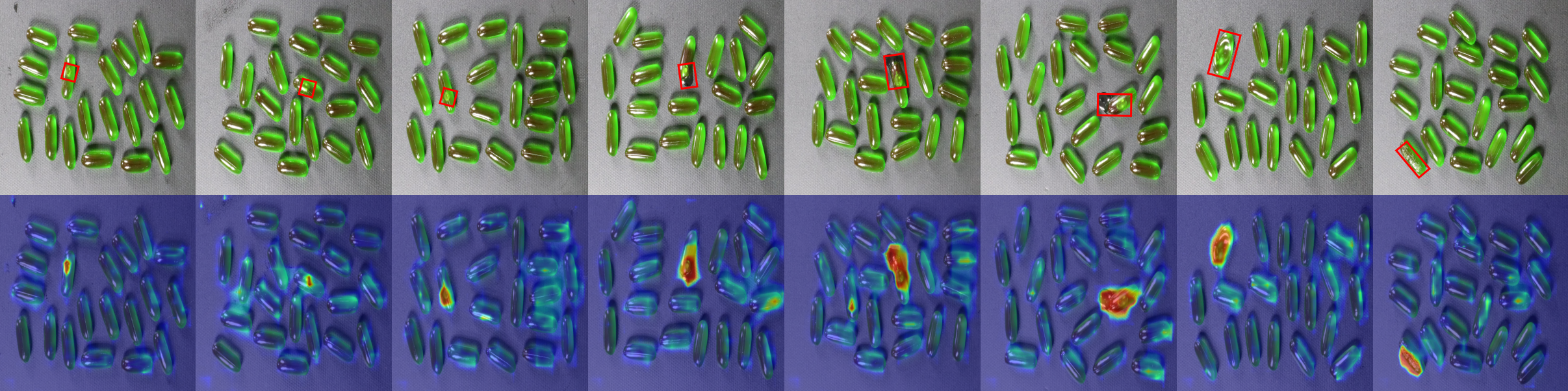}
  \caption{This visualization showcases the capsule product from the VisA dataset. The first row represents the input with the anomaly highlighted. The second row presents the segmentation results
from our model.}
   \label{fig:capsules}
\end{figure*}

\begin{figure*}
  \centering
  \includegraphics[width=\textwidth]{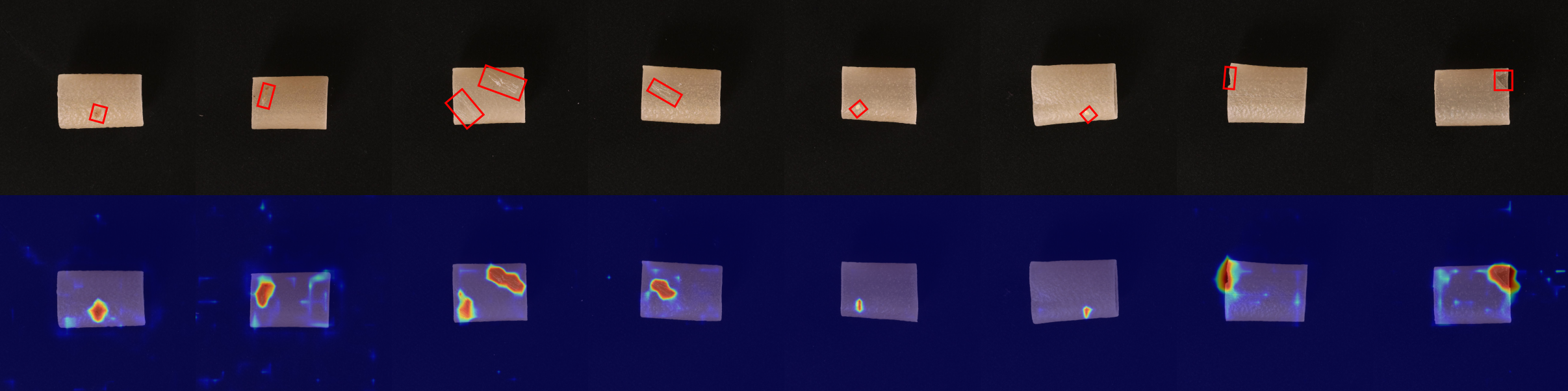}
  \caption{This visualization showcases the pipe fryum product from the VisA dataset. The first row represents the input with the anomaly highlighted. The second row presents the segmentation results
from our model.}
   \label{fig:pipe-fryum}
\end{figure*}

\begin{figure*}
  \centering
  \includegraphics[width=\textwidth]{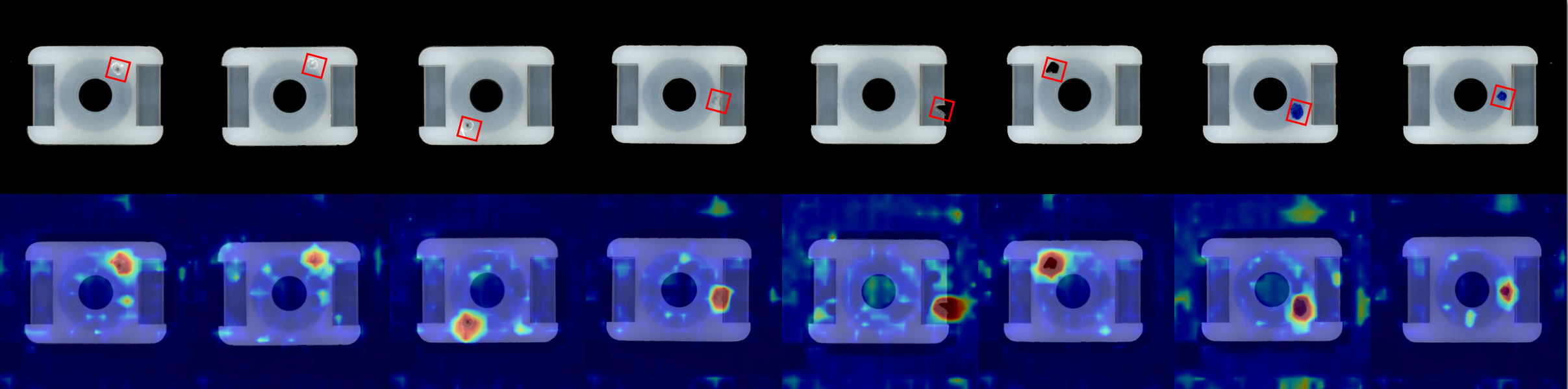}
  \caption{This visualization showcases the mount product from the Real-IAD dataset. The first row represents the input with the anomaly highlighted. The second row presents the segmentation results
from our model.}
   \label{fig:mounts}
\end{figure*}

\begin{figure*}
  \centering
  \includegraphics[width=\textwidth]{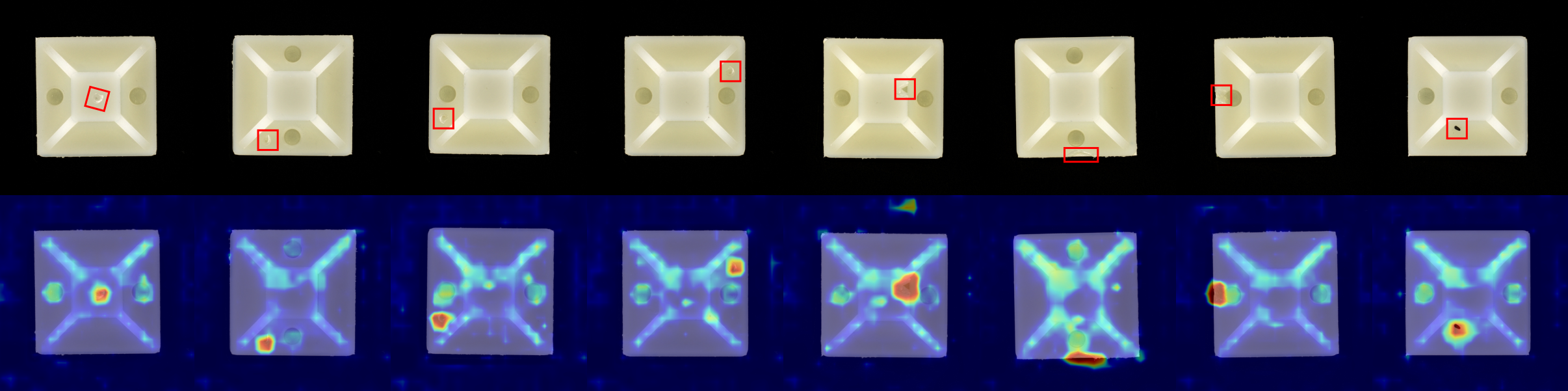}
  \caption{This visualization showcases the rolled strip base product from the Real-IAD dataset. The first row represents the input with the anomaly highlighted. The second row presents the segmentation results
from our model.}
   \label{fig:rolled_strip_base}
\end{figure*}

In this section, we provide additional visualizations of anomaly segmentation results together with example text responses generated by the model. We present eight samples from diverse products, including tubes (\cref{fig:tubes}) from MPDD, capsules (\cref{fig:capsules}) and pipe fryum (\cref{fig:pipe-fryum}) from VisA, and mounts (\cref{fig:mounts}) and rolled strip base (\cref{fig:rolled_strip_base}) from Real-IAD.

For the tubes, the model precisely identifies the deformed section and preserves clean boundaries, introducing almost no spurious activations elsewhere in the images. A similar pattern holds for the capsules, where the model distinguishes among different defect manifestations, correctly localizing internal bubbles, subtle surface scratches, and areas showing signs of leakage. For the pipe fryum, the model maintains high confidence across several defect types, producing focused masks for color specks, fine scratches, and fractured edges, and doing so consistently across varying shapes and orientations.
On the Real-IAD mounts, the model is able to localize defects but introduces substantial prediction noise. It also occasionally assigns elevated similarity scores to background regions, even when those regions are uniform black and contain no structural cues. For the rolled strip base, the model effectively captures the primary defect, although the anomaly map exhibits considerable background noise and incorrectly attributes medium similarity scores to structural features such as the central diagonal ribs, despite these regions being non-defective.

\begin{figure*}
  \centering
  \includegraphics[width=\textwidth]{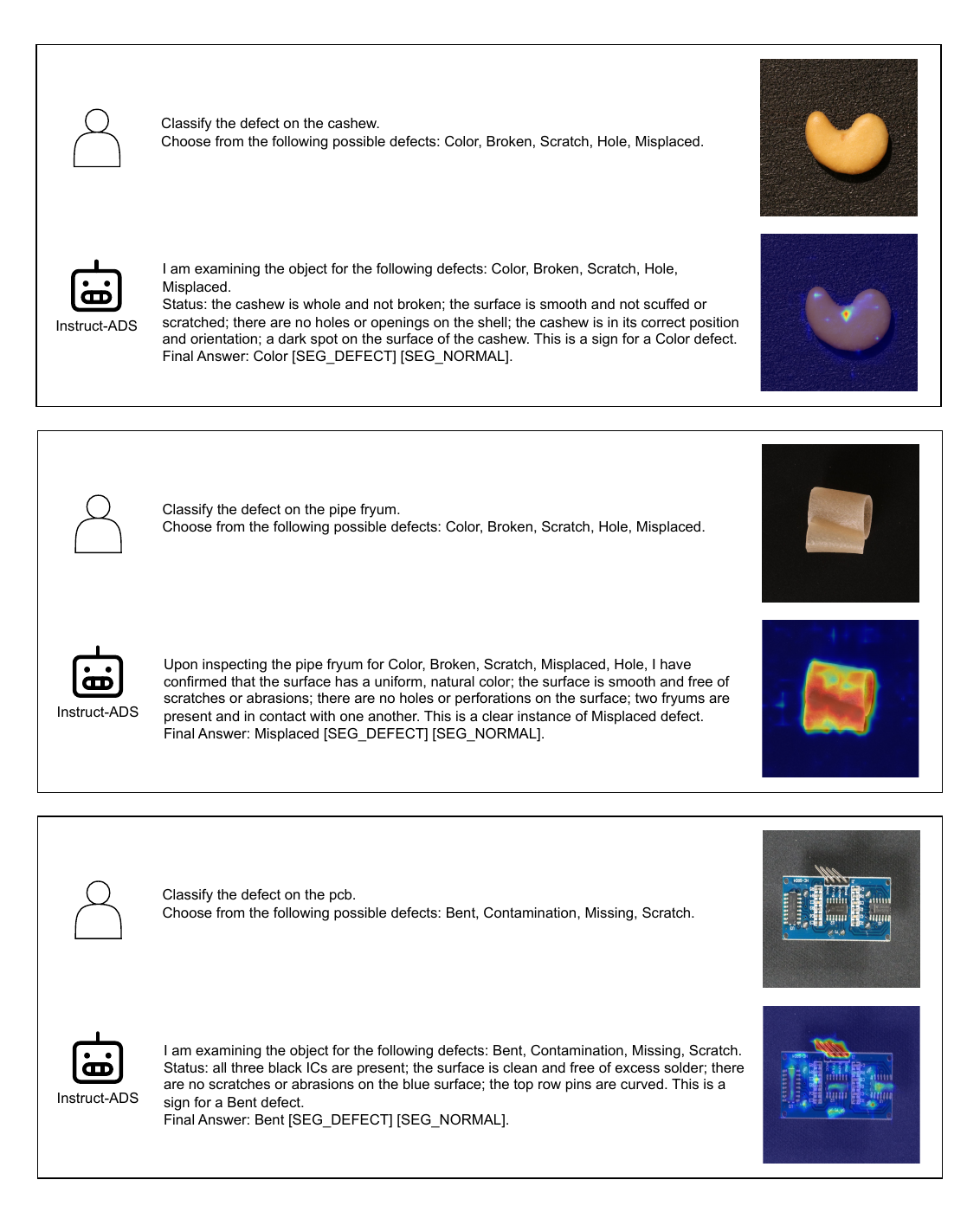}
  \caption{This visualization showcases the abilities of GenAU. Given the input prompt and the image, the model reasons over the possible defects, identifies the correct one and outputs a segmentation mask.}
   \label{fig:text}
\end{figure*}

In \cref{fig:text}, we show examples of how the model classifies defects across different products together with the corresponding segmentation masks. For the cashew sample, it correctly notes the presence of a dark region on the surface, corresponding to a \textit{color} defect. For the pipe fryum, it identifies that several pieces are touching and classifies the issue as \textit{misplaced}. For the PCB, it detects the \textit{bent} defect by recognizing the curved pins. Combined with the segmentation masks, these textual explanations provide detailed and actionable information that can support process optimization.

\end{document}